\documentclass{article}
\usepackage{tencent_ailab_tech_report}

\usepackage{amsmath}
\usepackage{graphicx}
\usepackage[colorlinks=true, allcolors=blue]{hyperref}

\usepackage{multirow}
\usepackage{subfigure}
\usepackage[utf8]{inputenc} 
\usepackage[T1]{fontenc}    
\usepackage{hyperref}       
\usepackage{url}            
\usepackage{booktabs}       
\usepackage{amsfonts}       
\usepackage{nicefrac}       
\usepackage{microtype}      
\usepackage{soul}
\usepackage{multirow}
\usepackage{graphicx}       
\usepackage{makecell}
\usepackage{hyperref}
\usepackage{wrapfig}
\usepackage{marvosym}
\usepackage[shortlabels]{enumitem}
\definecolor{myblue}{RGB}{215,238,247}
\definecolor{mygreen}{RGB}{230,241,221}
\definecolor{mygrey}{RGB}{242,242,242}
\definecolor{myorange}{RGB}{235,142,71}
\definecolor{textgreen}{RGB}{135,201,195}
\definecolor{mypurple}{RGB}{222,213,255}
\definecolor{darkblue}{RGB}{66,141,191}
\definecolor{greyblue}{RGB}{208,220,232}

\def\AM{{\mathcal A}}

\def\MM{{\mathcal M}}

\def\SM{{\mathcal S}}

\def\Name{Cognitive Kernel}

\title{Cognitive \ Kernel: \ An \ Open-source \ Agent \ System \\ towards \ Generalist \ Autopilots}

\author{Hongming Zhang$^\ddagger$, Xiaoman Pan, Hongwei Wang, Kaixin Ma, Wenhao Yu, Dong Yu\\Cognitive Kernel Team\thanks{
All authors contribute equally and are listed randomly. $^\ddagger$Hongming Zhang is the project lead. 
}, \ Tencent AI Lab, Seattle\\
\textit{\{hongmzhang, xiaomanpan, hongweiw, kaixinma, wenhaowyu, dyu\}@global.tencent.com}}

\begin{document}
\maketitle

\begin{abstract}

We introduce \Name, an open-source agent system towards the goal of generalist autopilots. 
Unlike copilot systems, which primarily rely on users to provide essential state information (e.g., task descriptions) and assist users by answering questions or auto-completing contents, autopilot systems must complete tasks from start to finish independently, which requires the system to acquire the state information from the environments actively.
To achieve this, an autopilot system should be capable of understanding user intents, actively gathering necessary information from various real-world sources, and making wise decisions.
\Name~adopts a model-centric design. In our implementation, the central policy model (a fine-tuned LLM) initiates interactions with the environment using a combination of atomic actions, such as opening files, clicking buttons, saving intermediate results to memory, or calling the LLM itself. This differs from the widely used environment-centric design, where a task-specific environment with predefined actions is fixed, and the policy model is limited to selecting the correct action from a given set of options. 
Our design facilitates seamless information flow across various sources and provides greater flexibility.
We evaluate our system in three use cases: real-time information management, private information management, and long-term memory management. 
The results demonstrate that \Name~achieves better or comparable performance to other closed-source systems in these scenarios.
\Name~is fully dockerized, ensuring everyone can deploy it privately and securely.
We open-source the system and the backbone model to encourage further research on LLM-driven autopilot systems\footnote{\url{https://github.com/Tencent/CogKernel}}.

\end{abstract}

\section{Introduction}\label{sec:introduction}


Large language models (LLMs) have revolutionized the landscape of AI applications~\citep{achiam2023gpt,team2023gemini,zhao2023survey,chang2024survey}. 
Systems like ChatGPT All-in-One\footnote{\url{https://chatgpt.com/}} and Microsoft's Copilot\footnote{\url{https://copilot.microsoft.com/}}, representing chat models and auto-completion tools respectively, have significantly enhanced productivity in everyday tasks.
However, these systems primarily function as ``Copilots,'' where users are still required to manage the majority of the work, such as planning the overall workflow, posing the right questions, or refining the model's output as needed~\citep{xi2023rise,wang2024survey}.
To fully harness the capabilities of LLMs and reduce the burden of tedious, repetitive tasks, we shall shift from building ``Copilot'' to ``Autopilot'' systems that can independently complete tasks.
For instance, while a Copilot system might assist in drafting a template for an invitation email, an Autopilot system should be capable of composing the entire email and sending it autonomously.
 




Recently, significant efforts have been directed toward developing agent systems to achieve autopilot capabilities~\citep{DBLP:journals/fcsc/WangMFZYZCTCLZWW24}.
For each specific task, such as debugging~\citep{DBLP:journals/corr/abs-2310-06770}, the agent system relies on a task-dependent environment to generate prompts~\citep{yang2024swe,he2024webvoyager,xie2024osworldbenchmarkingmultimodalagents,DBLP:journals/corr/abs-2307-13854,xi2024agentgym}.
These prompts typically include a description of the current environment and a predefined set of actions that the policy model (i.e., the LLM in this context) can choose from.
The policy model then selects the appropriate actions, which are sent back to the environment for parsing and execution, followed by waiting for the updated states.
This environment-centric design paradigm simplifies the task for the policy model, enabling the development of powerful, task-specific agent systems using static, closed-source LLMs like GPT-4 \citep{zhang2024autocoderover,guo2024dsagent,zhang2024webpilot}.
However, this paradigm is insufficient to support general-purpose tasks due to the vast state and action spaces involved.




To advance closer to the goal of a generalist autopilot system, we propose a model-centric design rather than an environment-centric approach.
Specifically, upon receiving a task, the policy model will generate step-by-step plans, execute corresponding actions, and actively gather new state information as needed.
This enables the model to dynamically adjust its strategies and enhance its adaptability in unforeseen situations.
This approach contrasts with traditional methods where the model passively waits for state descriptions from task-specific environments.

\begin{figure}
    \centering
    \includegraphics[width=\linewidth]{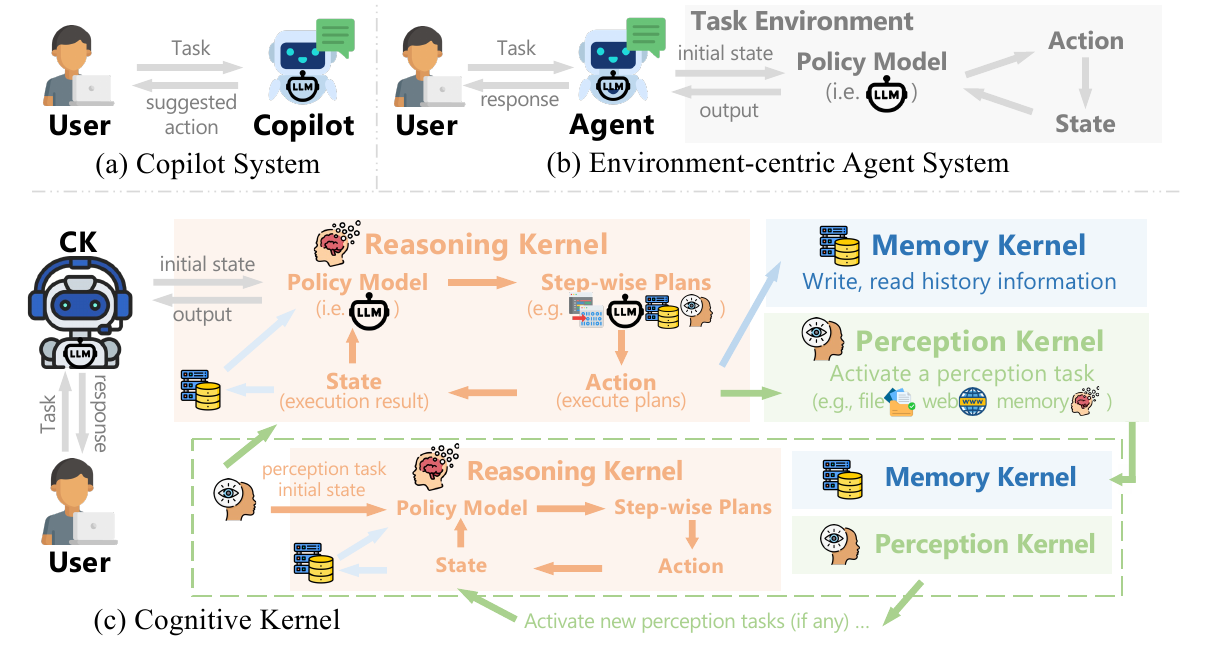}
    \caption{Comparison of conceptual frameworks: (a) Copilot system, (b) Environment-centric Agent system, and (c) \Name~system, highlighting key structural differences. 
    After receiving a task, \Name~will evaluate whether it has all essential state information to make a sound action.
    If not, it will actively perceive the missing state information from the environment, which can be a deeper-level self-contained autopilot task.
    }
    \label{fig:framework}
\end{figure}

In line with this design principle, we developed \Name, a system composed of three conceptual components: the reasoning kernel, the perception kernel, and the memory kernel, as illustrated in \autoref{fig:framework}(c).
These components correspond to decision-making, state perception, and state storage, respectively.
Analogous to a Turing Machine \citep{turing1936computable}, the reasoning kernel functions as the transition mapping mechanism, the perception kernel serves as the tool for reading current state information, and the memory kernel acts as the tape, recording past state information.
As a model-centric system, at any given step (including both the initial and intermediate steps informed by the perception kernel), the reasoning kernel — essentially an LLM-based policy model\footnote{Throughout this technical report, we name the LLM employed by \Name~as the ``policy model.''} — generates the next action.
This may involve activating the perception kernel to gather missing state information or engaging the memory kernel to store or retrieve critical historical information.
To mimic how humans acquire information during daily tasks and enhance generalizability, we encourage the reasoning kernel to utilize atomic actions — such as opening a file or clicking a button—that an ordinary person could perform, rather than relying on predefined high-level actions like API calls.
The system autonomously combines these atomic actions into compound operations for more complex tasks.
To facilitate this process, we choose a programming language (i.e., Python) rather than plain text as the medium for generating actions, due to its capability in defining compound actions and its efficiency in executing recursive tasks.
As depicted in Figure \ref{fig:framework}, perceiving the state itself is treated as a fine-grained autopilot task, potentially triggering the reasoning and memory kernels again. Consequently, these three conceptual kernels are tightly interwoven.
To enhance the system's efficiency and robustness, we reorganize these conceptual kernels into several separate Docker containers, with further details provided in Section \ref{sec:system_design}.

Switching from an environment-centric to a model-centric paradigm significantly enhances a system's generalizability but introduces new challenges for the policy model, particularly for closed-source models.
In an environment-centric design, prompts can be meticulously crafted to be self-contained, which aligns with the current training paradigm of LLMs, allowing for optimal performance in specific environments.
However, in a model-centric paradigm, the input prompts may no longer be self-contained, and the model must interact with the environment independently, leading to suboptimal performance in closed-source LLMs.
Additionally, using a fixed model limits the system's ability to continuously improve itself through interacting with users.
As a result, directly applying a powerful closed-source language model, such as GPT-4, may yield unsatisfactory performance.
To address this issue, \Name~employs a two-stage training process that enables the policy model to develop fundamental capabilities and learn from interaction feedback for further improvement.
In the released version of \Name, we initialize the system with the open-source model LLama3 \citep{dubey2024llama} and fine-tune it using a combination of open-source data and data collected through the system itself to better adapt to the system's design.

We evaluate the system's performance across three core capabilities that we consider essential for an ``autopilot'' system: real-time information management (e.g., gathering data and completing tasks on the open web), private information management (e.g., processing and understanding local files), and long-term memory management (e.g., personalizing the system based on user interactions).
Unlike existing works that primarily assess model capabilities, our evaluation emphasizes the entire system, including both the system design and the underlying model.
By comparing our system with leading closed-source systems (e.g., ChatGPT website, Kimi website) and various open-source agent systems utilizing different foundation models, we have made the following observations:
\begin{enumerate}[leftmargin=*]
    \item No single system consistently delivers the best performance across all tasks; each system has its strengths as well as inherent behavioral biases.
    \item Optimal performance is achieved through the deep integration of model and system design. \Name~demonstrates significantly better results when paired with our specifically adapted model compared to an overall stronger model like GPT-4o.
\end{enumerate}

To foster future research in the ``autopilot'' domain, we release the source code of the whole system and model weights, enabling everyone to deploy and use \Name~locally.
We hope \Name~can serve as an initial prototype of an ``autopilot'' system, inspiring the development of more advanced systems in this area.
It is important to note that while \Name~incorporates several internal engineering safeguards to prevent harmful actions, an ``autopilot'' system inherently possesses the capability to interact with and alter the real world.
Therefore, we strongly recommend that users employ \Name~solely for research and at their own risk.

The rest of this paper is organized as follows.
Section \ref{sec:backgrounds} provides background information on automated systems, offering a clear roadmap of prior efforts from both technical and user perspectives, particularly about agents across various literatures.
Section \ref{sec:theoritical_formulation} presents a formalized notation of a model-centric agent system.
Following that, Section \ref{sec:system_design} delves into the system design and training details, while Section \ref{sec:evaluation} thoroughly evaluates its performance in comparison to leading agent systems on realistic benchmarks. Section \ref{sec:applications} demonstrates \Name's capabilities through case studies that highlight tasks unachievable by existing systems, showing how \Name~can assist in daily work.
Finally, Section \ref{sec:conlusion} concludes this technical report.

\section{Backgrounds}\label{sec:backgrounds}

We first introduce previous efforts on rule-based automated systems, which motivates our formulation of an ``autopilot'' system.
Then we cover more recent works in building and training task-specific agent systems.

\subsection{From Rule-based Automated Systems to LLM-driven ``Autopilots''}
One of the automated system pioneers is the Turing machine~\citep{turing1936computable}, a mathematical model describing an abstract machine that manipulates symbols based on a table of rules and a foundation of modern computers. 
A Turing Machine contains two key concepts: states, the set of predefined variables, and transition functions, which are predefined rules describing how a system shall respond to a state.
Although traditional Turing machines cannot directly serve as an ``autopilot'' system to solve real-world tasks due to the almost infinite space of possible states and transition rules with randomness among them, it motivates the two fundamental duties of an ``autopilot'' system: (1) perceiving and managing essential states; (2) make wise decisions based on the states.

By compressing and modeling vast amounts of world information, LLMs implicitly learn the possible connection between states. They thus can serve as a powerful approximation of the transition functions and partially solve the second duty \citep{wu2024stateflow,ma2023laser}.
However, how to efficiently and accurately perceive and manage the state remains unclear.
Daily applications require both global state information (e.g., world knowledge) 
and localized state information that changes constantly (e.g., update-to-date knowledge or private information).
LLMs can learn to model the global state information but cannot model the localized ones.
Thus, how to manage the localized state information becomes the critical design choice for building an LLM-driven AI system.
For a Copilot system, we leave the task of providing localized state information to users, which simplifies the task but also limits the capability of the system to solve tasks independently \citep{chen2021evaluating}.
In contrast, for an autopilot system, we expect it to monitor and acquire the localized state information by itself and thus it has the potential to solve a complete task without human involvement \citep{significant_gravitas_2024}.
Therefore, in \Name~, we specifically designed the perception kernel and the memory kernel to perceive and memorize localized state information, and their orchestration is completely handled by the reasoning kernel, moving one step closer to an LLM-driven ``Autopilot'' system.

\subsection{Recent Advancement in Model-based Agents}

The concept of an agent, popular in the reinforcement learning (RL) community, has been foundational to artificial intelligence~\citep{watkins1992q,kaelbling1996reinforcement}. 
An agent interacts with an environment to learn how to achieve a specific goal by maximizing cumulative rewards~\citep{arulkumaran2017deep,sutton2018reinforcement}. Early work focused on the theoretical underpinnings of RL, such as Markov decision processes and dynamic programming~\citep{howard1960dynamic}. These methods provided the basis for the development of various RL algorithms. Recent advances like Q-learning~\citep{watkins1992q} and policy gradients~\citep{williams1992simple,sutton1999policy} have been crucial, especially with the introduction of deep RL, which combines neural networks for approximating Q-values, as seen in Deep Q-Networks~\citep{van2016deep} achieving human-level performance in complex tasks. Key algorithms such as DDPG~\citep{lillicrap2015continuous} and PPO~\citep{schulman2017proximal} have significantly impacted robotics, healthcare, finance, gaming, and autonomous driving, showcasing RL's broad applicability~\citep{gottesman2019guidelines,yu2021reinforcement,charpentier2021reinforcement,silver2018general,sallab2017deep,kiran2021deep}.


However, previous assumptions in RL significantly differ from human learning processes, as the human mind is highly complex and capable of learning from a much wider variety of environments.
With the recent developments of large language models (LLMs), the concept of agents has evolved beyond simple policy functions in restricted environments~\citep{xi2023rise,wang2024survey}. 
LLM-based agents can possess more comprehensive world knowledge, perform more informed actions, and also provide natural language interfaces for human interaction, making them more flexible and explainable~\citep{yaoreact,liu2023bolaa,he2024webvoyager,gur2023real,yang2024swe}.
For example, ReAct~\citep{yaoreact} prompts LLMs to perform dynamic reasoning to create, maintain, and adjust high-level action plans (reason to act), while also interacting with external environments (e.g., Wikipedia) to incorporate additional information into reasoning (act to reason), thereby achieving superior performance in benchmarks. However, prompting-based agent frameworks still perform poorly in many real-world agent scenarios. 

To make LLM-agents task experts, a series of works have focused on collecting expert trajectories from diverse environments and tasks, and training LLM-based agents through behavioral cloning~\citep{zeng2023agenttuning,chen2023fireact}. However, obtaining these expert trajectories is often costly and lacks sufficient exploration.
Another line of work involves training LLM-based agents based on environmental feedback using RL methods to align LLMs with agent task objectives~\citep{christianos2023pangu}. Additionally, some approaches utilize self-improvement, where the model explores the environment to obtain high-reward trajectories and fine-tunes itself based on these trajectories \citep{putta2024agent}. Although these training strategies have shown promising performance in reasoning, coding, and web tasks, these trained agents remain constrained to their specific tasks, struggle to generalize to general-purpose usage, and are restricted to pre-defined environments.

Existing efforts on LLM-driven agents often follow an environment-centric design, where they partially solve this problem by designing an environment for each task to manage and feed localized state information \citep{liu2023agentbench,xi2024agentgym}.
Moving one step further and aiming to create a general-purpose ``autopilot'' system that has the potential to solve more general tasks, \Name~switchs from environment-centric to model-centric and asks the system to actively perceive the localized state information with tools an ordinary person could use.



\section{A Conceptual Framework of ``Autopilot'' Systems}\label{sec:theoritical_formulation}

Motivated by previous automated systems~\citep{turing1936computable},
an autopilot system $AS$ should excel at managing the current state and making wise actions accordingly. 
Thus, we first formulate the conceptual autopilot framework as a 6-tuple $AS = \langle \SM, s_{n}, \AM, a_{n},  T, \MM \rangle$, where $\SM$ is the set of all possible states, $s_{n} \in \SM$ is the state\footnote{Each state is a set of variables and their values. In this paper, we omit the concept of variables for simplicity.} at timestamp $n$, $\AM$ is the set of possible actions, $a_{n} \in \AM$ is the action at timestamp $n$, $T$ is the transition matrix, which determines $a_{n}$ based on $s_{n}$, $\MM$ is the memory component that records $s_{0}$ to $s_{n-1}$.
Note that in real applications, $\SM$ and $\AM$ are arbitrarily large to be enumerated by modern machines.
As the size of $T$ is $|\SM| \cdot |\AM|$, $T$ is also too large to be enumerated.


To address these limitations and create a practical autopilot system, we further decouple the states into global and localized ones, where the global state information is the world knowledge shared by most humans and localized state information is temporally or spatially unique to the current task.
Specifically, we can decouple $\SM$ as:
\begin{equation}
    \SM = \SM^{g} \cup \SM^{l},
\end{equation}
where $\SM^{g}$ and $\SM^{l}$ represent the set of global and localized state information, respectively.
Similarly, for any state $s_n$ at timestamp $n$, we can also decouple it as:
\begin{equation}
    s_n = s_n^{g} \cup s_n^{l}.
\end{equation}

Based on the assumption that large language models such as GPT-4 have compressed the world's knowledge through the pre-training, we can use an LLM as the policy function $F$ to simulate $\SM^{g}$, $s_n^{g}$, and $T$.
Thus, we can reformulate $AS$ as $\langle \SM^{l}, s_{n}^{l} \AM, a_{n},  F, \MM \rangle$, where $F$ is the LLM-based policy function that 
predicts $a_{n}$ conditional on $s_{n}$.
Since LLMs are essentially probabilistic models, this formulation no longer guarantees the execution correctness, and thus $a_{n}$ can only be viewed as the most likely action based on the trajectories that $F$ has seen during the training phase.

A remaining challenge is where to get the localized state information in a real system. As discussed in Section~\ref{sec:introduction}, a key difference between autopilot systems and copilot systems is that the state information might not be provided and the system must perceive the state information actively.
To better represent this, we add one more variable $s_n^{o} \subseteq s_n^{l}$ to denote the observed localized states at each step $n$.
Empirically, $s_n^{l}$ is the optimal local state that one can not easily get, and $s_n^{o}$ is the localized state information the autopilot system truly has and can rely on.
At each step $n$, $F$ will first determine whether $s_n^{o}$ is close enough to $s_n^{l}$, if it is not close enough, $F$ will initiate a perception task to perceive more localized states.
Since each perception task might be an autopilot task requiring further state perception and planning, we denote it as $P^k$, where $k$ is the depth of the perception tasks.
Thus, we can get the final formal formulation of $AS$ as:
\begin{equation}\label{eq:formulation}
\begin{aligned}
    AS &= \langle \SM^{l}, s_n^{l}, s_n^{o}, \AM, a_{n},  F, \MM, P^0 \rangle, \\
    P^k &= \langle \SM^{l}, s_n^{l}, s_n^{o}, \AM, a_{n},  F, \MM, P^{k+1} \rangle.
\end{aligned}
\end{equation}
In Section~\ref{sec:system_design}, we will cover the implementation details of \Name~and explain how it fulfills the above formulation.



\section{System Architecture and Implementation}\label{sec:system_design}

As introduced in Section~\ref{sec:introduction}, \Name~contains three conceptual components: reasoning kernel, perception kernel, and memory kernel, which handles the three duties of an autopilot system: predicting the next action $a_n$ based on current observed state $s_n^o$, tracking the current state with perception tasks $P$, and storing and retrieving the past states with the memory component $\MM$, respectively.
In this section, we first introduce the motivations and design principles of the three kernels.
After that, we introduce the dockerized implementation details.
In the end, we introduce the training details of the center language model, which serves as the policy function $F$.

\subsection{Reasoning Kernel}
The reasoning kernel is responsible for generating a plan for the next moves and then executing it.
Creating a general-purpose autopilot system has many practical challenges.
The first one is uncertainty. 
The real world is full of uncertainty. 
Even with all the essential state information, a model still cannot accurately predict the effect of its action. Moreover, complex tasks usually involve long trajectories of actions to complete, leading to exponential growth in the degree of uncertainty. Thus, it's almost impossible for the system to generate a perfect plan that can execute end-to-end in the beginning. 
The second challenge is the efficiency. 
It is widely known that existing language models are huge and the inference can be time-consuming.
In an LLM-driven agent system, since every step is an inference task that takes considerable time even with the latest inference frameworks, the task completion process could be slow and thus provide a bad user experience. 

To address these challenges, we follow existing works to use the programming language (i.e., Python) as the medium for planning and execution \citep{li2023chain,zhang-etal-2024-natural}, where the basic Python operations such as addition are considered to be the atomic actions and Python functions are considered to be the compound actions. 
Compared with natural language, the programming language provides great flexibility for handling uncertainty. For example, the policy model could use ``if/else'' statements to design alternative strategies for task completion or use ``for loop'' to iteratively attempt different options. This enumeration operation is typically infeasible for natural language. 
Also, programming language provides a much higher level of parallelism than natural language, allowing multiple steps to be executed simultaneously. For example, for a task that requires checking information from multiple sources to cross-validate, a natural language-based agent system has to check each source iteratively. But with the programming language, one could easily speed up this process by running a piece of multi-threading code to check all sources concurrently.

To fully unleash the power of PL-centric planning, we further make a few improvements over the past code-centric agent systems. 
Firstly, instead of single-time execution, \Name~implements a state caching mechanism to cache previous execution states and functions for future usage.
As a result, if the later planning steps require the intermediate results from previous states, it can use the cached states without repeatedly generating and executing the past code. This allows the system to generate a partial plan, execute it, and then decide the next moves based on the results and so on. Hence the system can account for the uncertainty and run efficiently at the same time. 
Secondly, we implement an async parallel execution mechanism in \Name, enabling independent steps to run simultaneously and avoiding the slow actions blocking the whole execution process.




\subsection{Perception Kernel}

The perception kernel is responsible for accessing the environment, which is the real world in our scenario, to perceive the localized states.
Actively activating the perception kernel to get localized state information is the core functionality of an autopilot system. 
This subsection introduces how \Name~perceives two kinds of localized state information.
The first category is temporally localized information, which refers to the information that is constantly changing such as the weather information or the opening hours of specific restaurants.
The second category is spatially localized information, which refers to the information that can only be accessed by the local user and is not available anywhere else.

\subsubsection{Temporally Localized State Perception}\label{sec:web}

The world is constantly changing and the autopilot needs to access the up-to-date information.
For humans, the easiest way is using the internet.
Similarly, we also equip \Name~with the access to the open web.
Existing systems such as ChatGPT All-in-one, Gemini, or KimiChat leverage search engines to gather updated real-world information. However, this approach is inherently limited because it cannot perform more complex interactions with different websites. Instead, we give the system more freedom and allow it to directly control a live browser to interact with the open web like a person, e.g., by performing atomic actions such as clicking and typing.
By doing so, \Name~could finish more complex tasks such as ``finding the latest commit details of a popular GitHub repository'' that one cannot find answers from the search engine. 

After receiving a command from the user, the reasoning kernel will first generate a plan based on the current system status and the incoming user query. 
If the reasoning kernel thinks the system needs to perceive more temporal state information, it sends a request to the perception kernel with a specific instruction,  
which can be a more fine-grained level autopilot task as shown in Figure~\ref{fig:framework}.

After the perception kernel receives the command, it activates a web server\footnote{We implement the web server with the javascript version of the playwright for efficient and robust web browsing. More details can be found at \url{https://playwright.dev}.} to complete the task. 
At every step, the system will first observe the current web session as the state information and send that information to the reasoning kernel for the next action. 
We follow~\cite{DBLP:journals/corr/abs-2307-13854} and use the web page's accessibility tree as the observation to the agent due to its structural format and conciseness. We further optimized the raw accessibility tree to reduce redundancy and prune out irrelevant information. (See more details in Section \ref{sec:implementation_details})

Similar to how humans browse the web, we define the following atomic actions for \Name~to control the browser: (1) \textbf{Click}: click an element on the webpage; (2) \textbf{Type}: clear the text content in an element and fill it with new content; (3) \textbf{Scroll}: scroll up or down of the current viewport; (4) \textbf{Goback}: go back to the previously browsed page; (5) \textbf{Restart}: return to the homepage directly and restart the browsing process; (6) \textbf{Stop}: summarize the relevant information and sent back to the upper-level reasoning kernel.
At each step, after receiving the web description, \Name~will call the reasoning kernel to make a plan for the next step.
If the generated actions can be directly executed in the browser, \Name~ will execute it and provide an updated observation.
If an error occurs during action execution, the error message is also sent back to the reasoning kernel so it can revise the plan.  
The system continuously issues actions to explore the web until the stop action is generated or the trajectory reaches the maximum number of allowed steps.

\subsubsection{Spatially Localized State Perception}

Another perspective of the localized state information is the spatial one.
Real-world tasks often involve private information only accessible to the local system/user.
This section covers the two most popular localized state resources: local files and history.

\textbf{Files:} Files such as docs or spreadsheets play critical roles in information transferring in people's daily lives. 
Hence an ``autopilot'' system should also be able to perceive information from the local files to complete various tasks.
Similar to how humans process files, \Name~could use basic operations such as opening a file and searching for certain keywords by generating code in the reasoning kernel.
We classify these operations into four categories: (1) \textbf{Operate}: Perform specific operations such as counting occurrences, finding specific terms, and 
 extracting part of input data. (2) \textbf{Navigate}: Move to different file locations, such as the next page, the previous page, or a specific page number. (3) \textbf{Search*}: Perform semantic-based retrieval to find relevant information within the file\footnote{This operation requires the support of the memory kernel. For implementation efficiency, when \Name~receives the files, it will first add all contents into the memory kernel for later retrieval usage. More details will be covered by Section~\ref{sec:memory_kernel}.}. (4) \textbf{Read}: Understand and summarize the content, extracting key information, generating insights, or directly answering questions based on the file's content.
At each step, if \Name~thinks that the current state information is not enough and it needs to perceive more from the local files, it will activate the suitable operations, which can be the aforementioned atomic ones or the combination of them to perform more complex observations.

\textbf{Long-term History}: Besides files, another important spatially localized state information is the long-term history between users and the autopilot systems.
For example, one might expect his ``autopilot'' system to know the location of his home without mentioning it in every relevant command.
It is widely known that existing language models all have a limited context length, typically from two thousand to one million, and thus an autopilot system cannot store all past interactions with the user as the context.
To solve this problem, we treat history as a special kind of spatially localized information that is available to the current user and autopilot system and formulate the storage and usage of such information as a perception task. 
Specifically, if \Name~thinks that it needs to store some information in the history or perceive history, it will activate the memory kernel to write into or load from the memory just like how modern computers operate the disk. 
More technical details of our memory design will be covered in Section~\ref{sec:memory_kernel}.

    

\subsection{Memory Kernel}\label{sec:memory_kernel}

\begin{figure}
        \centering
        \includegraphics[width=\linewidth]{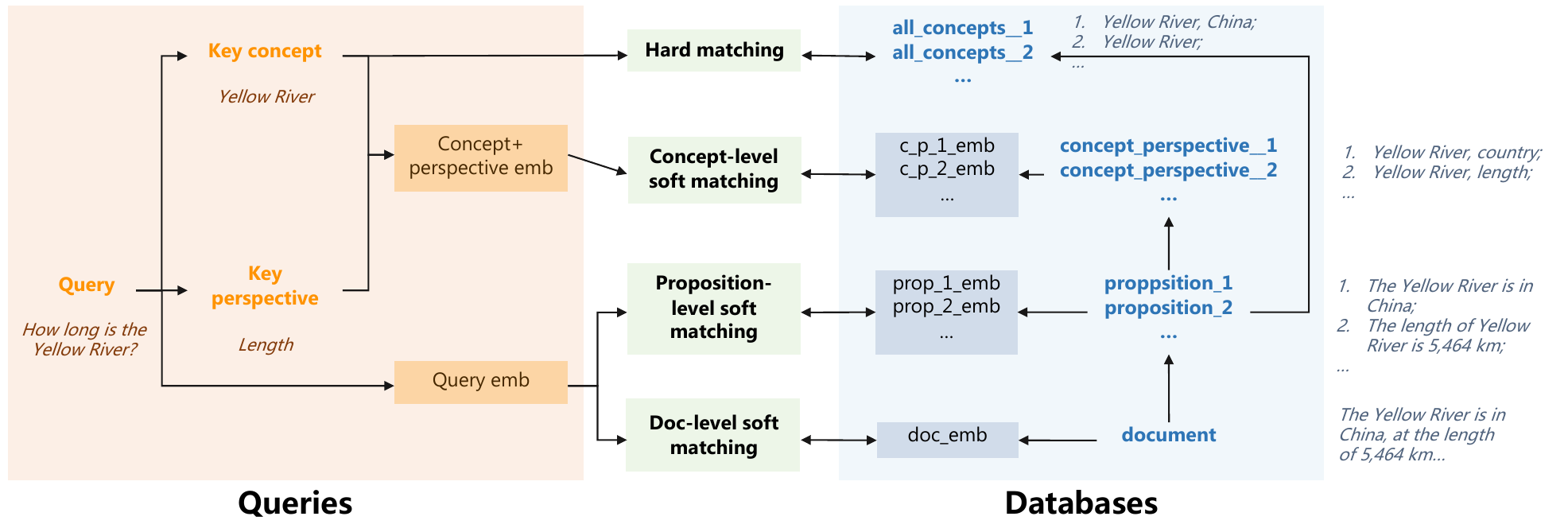}
        \caption{The overall framework of the multi-granularity information management system.}
        \label{fig:ir}
    \end{figure}

The main duty of the memory kernel is to provide a caching mechanism for the autopilot system to save and retrieve past states.
The most intuitive way of implementing such a module is using the widely used dense retrieval methods, which segment the content of interest into fixed-length chunks and create chunk indexes with representation models \citep{karpukhin2020dense, ni2021large, izacard2021unsupervised}.
However, this traditional approach is sub-optimal for an ``autopilot'' system.
The dense-retrieval model was originally trained to find relevant text pieces from a huge plain text database like Wikipedia. 
However, an ``autopilot'' system often needs to store and retrieve well-structured information that requires more fine-grained semantic matching.
To better suit this need, we propose a multi-granularity information management system as the memory kernel of \Name.
The overall framework is illustrated in \autoref{fig:ir}, which includes two major components: information processing/storage and information retrieval.

    


\textbf{Information processing/storage}.
The information processing/storage component is illustrated in the right part of \autoref{fig:ir} (the blue rectangle). For any given information, we first convert it into plain text and treat it as a regular document.
For example, for dialogue history with the timestamp information, we could create a sentence in the format of ``C@@T,'' where C and T represent the content and timestamp, respectively.
To help the system better retrieve such information, we parse the documents into different granularities and create semantics indexes (i.e., embeddings), accordingly.
Specifically, we consider the following granularities: (1) \textbf{Documents}: the most coarse one is the input document without any processing. 
We name the embedding representation of the input docs as (\textsf{doc\_emb}); (2) \textbf{Propositions}: A proposition is a semantically complete sentence without any compound structures~\citep{DBLP:journals/corr/abs-2312-06648}.  We break down all the sentences in a document into propositions, ensuring that the resulting propositions do not omit any content from the original text or add any extra content. 
For example, the document $d$ in the bottom right corner of \autoref{fig:ir} can be broken down into propositions of $p_1$ = ``\textsf{The Yellow River is in China},'' $p_2$ = ``\textsf{The length of Yellow River is 5,464 km},'' ...
We denote embedding representations of all propositions as (\textsf{prop\_1\_emb}, ...).
(3) \textbf{Key concept and perspective}: A finer granularity is the key concept and perspective in the propositions. In \Name, we extract each proposition's key concept and perspective. 
For example, the concept and perspective for $p_1$ is ``\textsf{Yellow River}'' and ``\textsf{country},'' and the concept and perspective for $p_2$ is ``\textsf{Yellow River}'' and ``\textsf{length}.''
To facilitate the semantic matching, we concatenate the concept and perspective as a phase and then compute the embedding representation (\textsf{c\_p\_1\_emb}, ...).
(4) \textbf{Mentioned Concepts}: Last but not least, we also keep all concepts in each proposition and use that for hard matching. For example, the mentioned concepts for $p_1$ are ``\textsf{Yellow River}'' and ``\textsf{China},'' and the mentioned concepts for $p_2$ are ``\textsf{Yellow River}.''

\textbf{Information retrieval}.
As shown in the left-side of Figure~\ref{fig:ir}, another job of the memory kernel is finding relevant information based on the input query given by the reasoning kernel.
Given any input query $q$, following~\cite{DBLP:journals/corr/abs-2312-06648}, we first compute the query representation (i.e., \textsf{Query\_emb}) and extract key concepts and perspectives with the same models we used for the processing step. 
And then, \Name~conducts the multi-granularity matching with the following granularities: (1) \textbf{document-level soft matching}: finds the most relevant documents based on the similarity between \textsf{Query\_emb} and \textsf{doc\_emb}; (2) \textbf{proposition-level soft matching}: finds the most relevant propositions based on the similarity between \textsf{Query\_emb} and \textsf{prop\_1\_emb} and then retrieves the corresponding documents; (3) \textbf{Concept-level soft matching}: finds the most relevant concept+proposition combinations based on the similarity between \textsf{Query\_emb} and \textsf{c\_p\_1\_emb} and then finds the corresponding proposition and documents; (4) \textbf{Concept-level Hard Matching}: finds all documents that share concepts with the input query.
Finally, \Name~reranks the retrieved documents based on their maximum similarity according to the above four matching methods.
For example, if the retrieved result of document-level soft matching is [(doc A, 0.8), (doc B, 0.7)] and the retrieved result of proposition-level soft matching is [(doc B, 0.9), (doc C, 0.6)], then the merged result of the two matching methods is [(doc B, 0.9), (doc A, 0.8), (doc C, 0.6)].

\subsection{Policy Model Training}
The duty of the policy model is to simulate the optimal transition matrix and make wise actions based on the current state.
Such actions include initiating interaction with the environment given incomplete state information, generating atomic actions for the perception kernel, and aggregating information from the memory and perception kernels into complete responses to users.
Since directly applying a closed-source model leads to unsatisfying performance, we trained our own model upon open-source language models (i.e., Llama3~\citep{dubey2024llama}).
The training contains two stages.
In stage one, we employ standard supervised fine-tuning to train our model. 
Specifically, we use a mixture of data including open-sourced instruction following data~\citep{zhou2023limaalignment,luo2023wizardcoder}, function calling data\footnote{\url{https://huggingface.co/datasets/glaiveai/glaive-function-calling-v2}}, agent trajectories data for various tasks \citep{zeng2023agenttuning,wang2024executable,yin2023agent,DBLP:journals/corr/abs-2307-13854}, and a small set of manually annotated data that fits our system design to train our model. 
This stage equips the model with the general problem-solving capability and the basic capability of invoking atomic actions defined in our system.
However, the output distribution remains relatively flat after the first training stage, leading to unstable performance, particularly with new inputs out of the training data distribution. 
To overcome this challenge and enhance the model's generalization ability, we conduct a second-stage training.
Specifically, we deploy the first-stage model online and then collect the system's output trajectories given various inputs. Again, we used a mixture of data where the inputs are either mined from open-source datasets \citep{wang2023openchat,he2024webvoyager,dasigi2021dataset,trivedi2022musique} or submitted by internal users. Here, to ensure that the data we collect are with high-quality, we also collect judgments and feedback for the system trajectories. The judgments and feedback can come from a user (where the user can directly submit via \Name~'s user interface), the system itself (when the code produces the error and error message), or an external model (e.g. GPT-4). We use the judgments to filter out bad cases and train the model on the successful trajectories to continue improving its ability. 
Furthermore, we can use this data to empower the policy model with the capability of criticizing whether the outputs fit the user's preference or not. 
More training details can be found in Appendix~\ref{app:training}.


\subsection{Implementation Details}\label{sec:implementation_details}

\begin{figure}
        \centering
        \includegraphics[width=\linewidth]{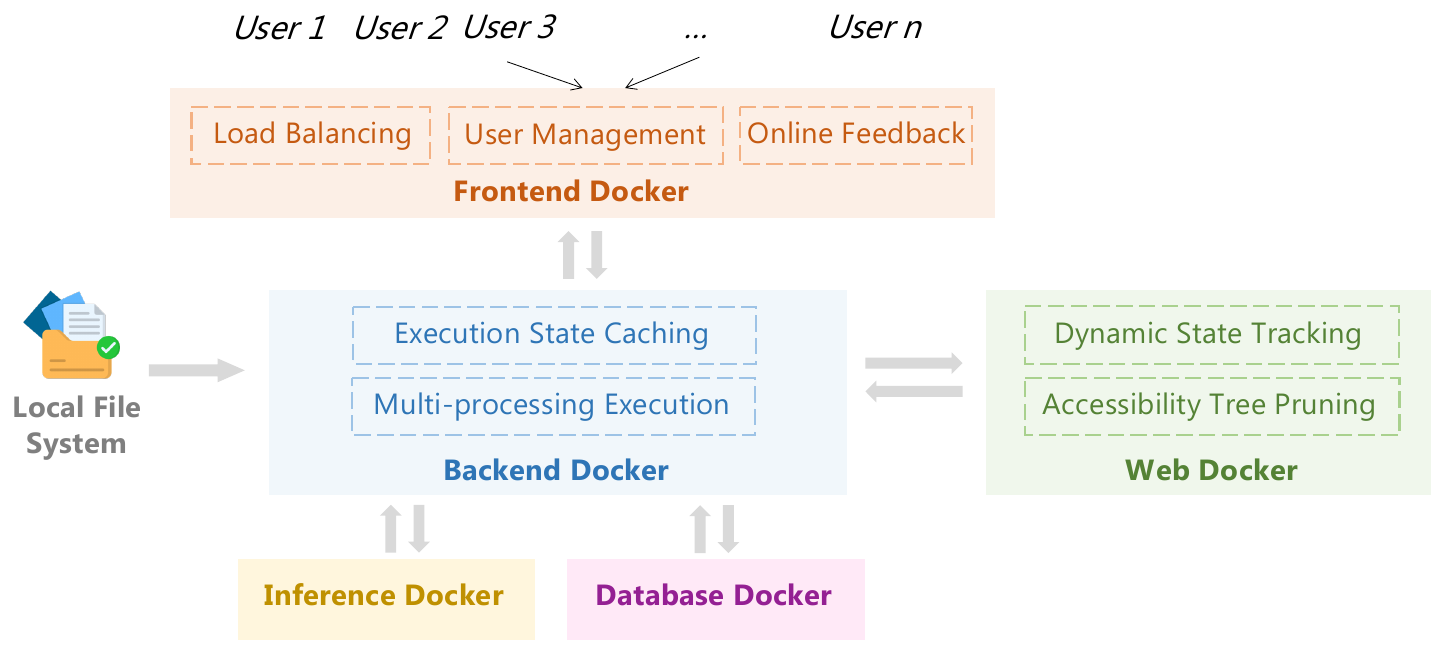}
        \caption{The engineering framework of \Name.}
        \label{fig:dockerized_design}
    \end{figure}

We adopted a dockerized design in our implementation to ensure efficient and safe deployment. As demonstrated in Figure~\ref{fig:framework} and Equation~\ref{eq:formulation}, the three conceptual kernels are deeply integrated.
Thus, for efficient scheduling and execution, we reorganize the system into separate dockers and optimize each one toward the assigned task.
As shown in Figure~\ref{fig:dockerized_design}, we implement \Name~as five separate dockers: (1) the frontend docker, which provides the user interface, supports interaction with multiple users and collects feedback for the continuous training; (2) the backend docker, which provides an isolated environment for planning execution and controls the workflow with all other dockers; (3) the web accessing docker, which provides an isolated environment for perceiving the localized state information via the open web; (4) the database docker, which handles the memory management; (5) the inference docker, which handles the inference of the central language model.
Dockers communicate with each other via APIs. 
This dockerized design guarantees \Name's high parallelism.
For example, the inference docker could batch all queries to the central policy model from different levels in Equation~\ref{eq:formulation} for more efficient inference.
Besides that, this dockerized design also provides an isolation environment for all modules and thus provides a safer and more robust system.
The implementation details are as follows.

\textbf{Frontend Docker:} An ``autopilot'' system should have an interface to receive user commands, demonstrate the progress, provide the execution results, and collect user feedback. An illustration of \Name~'s user interface is shown in \autoref{fig:UI} and an illustration of the feedback mode is in \autoref{fig:annotation}. We implement the frontend docker with the React package\footnote{\url{https://react.dev/}}. Considering that the system might be working with multiple users simultaneously, we use Nginx\footnote{\url{https://nginx.org/en/}} to balance the requests from different users. Since each user might have different preferences for \Name's behavior, we also implement a user management module to store personalized information so that other components could benefit from such information, e.g., personal preference databases. Essentially, every user can create an account with the system, and their chat sessions will be stored in an isolated database, which can also be used in the memory kernel. 
To effectively collect feedback from users and continuously improve the system performance, we implement an online feedback module, where the user can see all details of the system execution and provide comments or suggestions accordingly. We discuss more details about the frontend in Section \ref{app:frontend}. 

\textbf{Backend Docker:} The backend docker is the center of \Name. At each step, the backend docker converts the potentially incomplete state information $s^o$ into a prompt and sends it to the inference docker to generate plans for the next step, which is essentially the combination of text and a piece of code (unless the model determines that this is the final state). After that, the backend docker executes the planned actions. 
Since the code generation and execution can be time-consuming, to avoid the repetitive plan generation and execution, the backend docker includes an execution state caching mechanism.
Similar to a Jupyter Notebook\footnote{\url{https://jupyter.org/}}, for any request, the backend docker will cache all intermediate execution results including the variables and function definitions in the memory.
For later steps, the backend docker will include that information as part of the state description and thus the model can avoid generating repetitive code, which can significantly increase the planning generation efficiency.
Besides that, since the system only needs to execute the continued plans based on the cached states, we also save time on repetitive code execution on different steps.
Another optimization we did was the multi-processing synchronization. 
To increase the system efficiency, the backend docker adopts a multi-processing strategy to handle concurrent requests instead of processing the queue linearly.
One thing worth mentioning is that since these concurrent requests might use shared cached variables, we will create a copy of the reused variables and provide a divergence of the caching branches to avoid potential conflicts.

\textbf{Web Docker:} The web docker is responsible for maintaining the web browsers such that the system can get real-time information from the open web. 
Once the generated plan determines that the system needs to get the temporal localized state information from the web, the backend docker will send an instruction to the web docker, which will then activate a new/cached browser to complete the task.
To improve the system efficiency, we implement the web docker with the javascript version of Playwright\footnote{\url{https://playwright.dev/}}. 
Like the backend docker, we design the web docker to support multi-processing and execution state caching. 
Regarding the webpage observations, we found that directly providing the raw accessibility tree as the state description is sub-optimal due to the following challenges: 1) for long websites that contain tens of thousands of elements, the full accessibility tree could easily exceed the context length limitation of current models (e.g., 16K). 2) the accessibility tree might contain huge redundancy, for example, the compound web elements such as grid cells may repeat the same semantic information multiple times along the parent-children hierarchy. Such redundancy causes additional computation and may distract the model when predicting the actions. To address these challenges, the web docker first parallelly localizes the visible elements from the current viewport and only constructs the tree for these elements. 
Then, we perform node deduplication heuristically so that only the nodes containing unique semantic information are kept. 
Such a design significantly improves the system's efficiency while containing essential state information. 

For actions that require arguments (target element to interact), e.g., click and type, previous works typically rely on the element's coordinates within the viewport to execute the action \citep{DBLP:journals/corr/abs-2307-13854}. However, using the coordinates may lead to unexpected outcomes. For example, when dropdown menus or grid cells are expanded on certain websites, the coordinates of certain elements will overlap. 
As a result, if the system decides to click an element behind the dropdown menu, the execution will click on the elements in the dropdown menu, which is problematic.
To avoid such scenarios, \Name~uses element's role and name to pinpoint the target element to interact. This helps precisely execute the action regardless of the current layout of the elements on the webpage. 
Another challenge we are facing is the context length limitation of modern LLMs. 
Since most web browsing tasks require multiple atomic actions, the agent's trajectory could be long, which results in the overlength context.
To remedy this issue, we condense the agent's trajectory history by dynamically replacing previous observations with special tokens. 
Since historical actions often contain a thinking step before the real action and thus the modified trajectory can retain the relevant information from historical observations and better predict the current action. 

\textbf{Inference Docker:} The inference docker is responsible for receiving prompts from the backend docker and calling the central language model to generate plans for the next steps. In \Name, we support both TGI\footnote{\url{https://huggingface.co/docs/text-generation-inference/en/index}} and vLLM~\citep{kwon2023efficient} as the inference server.

\textbf{Database Docker:} The database docker provides support for all memory-relevant operations. We use multiple database systems to fit the different needs of \Name.
Specifically, for permanent information that we want to store for a long time 
such as user feedback, we use the postgresql\footnote{\url{https://www.postgresql.org/}} due to its high reliability.
For temporal content such as an uploaded file or cached execution results that are useful for the live sessions, we use sqlite\footnote{\url{https://www.sqlite.org/}} due to its lightness.
When dealing with user-uploaded files, the information \Name~perceives can be long and might easily exceed the context length of the center language model. To address this issue, we adopt a similar trick as in Web Docker that prioritizes retaining the generated plans at each step while summarizing or omitting redundant content from the observations. This approach ensures that the model retains relevant historical context without overwhelming the input sequence length.

\textbf{Local file system:} The local file system allows \Name~to access the localized information. For safety concerns, we only allow the backend docker to read from the local file and cannot write into it to minimize the potential risk of harmful code\footnote{Based on the actual deployment experience, normal usage will not cause any harm, this design is mainly used to protect the system from the intentional attack.}.




\section{Evaluation}\label{sec:evaluation}

We evaluate \Name~on tasks that best reflect how users would interact with the system when deployed.  
Specifically, we focus on evaluating \Name's ability to 1) gather real-time information and complete web-based tasks, 2) process user-uploaded files and answer questions, and 3) manage the interaction history with the user for better personalization. We mainly compare against the following general-purpose end-to-end AI systems in our experiments: \textbf{ChatGPT}\footnote{\url{https://chatgpt.com/}}~\citep{achiam2023gpt}, \textbf{Gemini}\footnote{\url{https://gemini.google.com/}}~\citep{team2023gemini}, \textbf{Claude}\footnote{\url{https://claude.ai/chat/}}, \textbf{Kimi}\footnote{\url{https://kimi.moonshot.cn/chat/}}, and \textbf{Coze}\footnote{\url{https://www.coze.com/}}, and we directly use their web interface for evaluation. We used GPT-4o for ChatGPT, Gemini-Pro-1.5 for Gemini, Claude-opus for Claude, and the default version for Kimi. For Coze, since it allows users to customize the bot, we used different configurations for different tasks for the best performance. In particular, we activated Browser, Google Web Search, and WebPilot plugins for web-based evaluation, Long-term memory for long-term memory evaluation, and Doc Reader for document-based evaluation. Finally, we also consider a baseline where we switch the backbone model of \Name~to GPT-4o to understand the impact of the central policy model. 
\subsection{Benchmarks}
For real-time information management evaluation, we conduct experiments on the recently released WebCanvas benchmark \citep{pan2024webcanvasbenchmarkingwebagents}. WebCanvas test set contains 104 human-annotated tasks that require interacting with real-world live websites to complete, and each of the tasks specifics a target website for interaction. We simply provide each task instruction to each system and it is expected to strictly follow the instructions to find the target webpage and complete the task. 

For private information management evaluation, we conducted extensive assessments using \textsc{DocBench}~\citep{zou2024docbench}. \textsc{DocBench} provides an end-to-end evaluation: starting with a raw file input along with user questions and evaluating the system based on the quality of the answers generated. This benchmark includes 229 real-world documents and 1,102 questions, spanning five distinct domains: Academia, Finance, Government, Law, and News. Furthermore, it encompasses four major types of questions: text-only, multi-modal, meta-data, and unanswerable questions. We simply upload the files to each system and ask the questions. 

For long-term memory management evaluation, since there is no existing benchmark to the best of our knowledge, we constructed an in-house test set.  
Specifically, each test case consists of a session of messages between the user and the assistant acting as dialog history, followed by a final question about the details of previous messages. The goal is to assess if the assistant can accurately retrieve the ground truth message(s) from the dialog history and correctly answer the query.
The dataset has four categories of questions:
Single Message (1-M), where answering the final question relies on a single ground truth message;
Multiple Messages (Mul-M), where answering the final question requires combining information from two or more ground truth messages;
Knowledge Update (Update), where the user initially provides some information and later updates or corrects it.
Temporal Reasoning (Temp), where the questions require inferring the chronological order of two events mentioned in conversation history.
We create the benchmark with both human written and LLM generation.
For the manually-created subset, we write both the dialog history and the final question and we only cover the first three categories.
For the LLM-generated subset, we prompt the LLM (i.e., GPT-4o) to first create users with particular attributes, preferences, and experiences, then we have an LLM role-plays the user while another LLM role-plays the assistant to create the dialog history.
Finally, an LLM is prompted to generate the question and the answer.
The generated conversations, questions, and answers are manually checked by humans to ensure quality. During the evaluation process, we manually enter the dialog history (user turns) to the tested systems, then start a new chat session and ask the final question.

\subsection{Metrics}
We mainly focus on the end-task success rate in our evaluation. However, the definition of task success and evaluation methods are different for each of the target scenarios. Here we provide the detailed definitions. 

For WebCanvas, \cite{pan2024webcanvasbenchmarkingwebagents} proposed step-wise scores that use human-annotated key nodes along the gold trajectory to evaluate the system's web browsing performance. 
Upon further inspection of the annotated key nodes, we found that this metric can significantly underestimate the system performance. Since there exist many possible trajectories that lead to task completion, simply matching the key nodes from one trajectory can overlook many other valid paths. Thus we opt for manual evaluation and we only focus on the overall task success rate. 
Here, since many tasks cannot be truly ``completed,'' we consider a task to be successful if 1) the required information is gathered from the target website and 2) all the necessary actions are performed with regard to the correct elements. For example, the system cannot truly buy a gift card (without valid account information), we consider it to be successful if it has added the correct gift card to the cart on the target website and filled in the fake user information in the instruction to the correct cells on the checkout page. We closely monitor the system's web interaction sessions during our experiments to ensure that no harm is done to website hosts. For systems that do not provide the intermediate trajectories, we consider a task to be successful if the system provides the links to the correct websites that satisfy all requirements in their responses. 

For \textsc{DocBench}, we follow \cite{zou2024docbench} and adopt GPT-4o to automatically evaluate the correctness of the generated answers based on the reference. As reported by \cite{zou2024docbench}, relying on string matching or number extraction to evaluate the accuracy of generated response can be imprecise, since different LLMs and systems exhibit substantial variations in the organization and style of their outputs, potentially leading to biases in traditional metrics.
We instruct GPT-4o to assign a score of 0 (incorrect) or 1 (correct), thus using Accuracy to measure system performance. 

For long-term memory evaluation, we ask Cognitive Kernel as well as baseline systems to generate the answer for each question, and manually judge the correctness of the answers.


\begin{table*}[t]
\centering
    \small
\setlength{\tabcolsep}{4pt}
\begin{tabular}{l|ccc}
    \toprule
    \textbf{Systems} & \textbf{Real-time Information} & \textbf{Private Information} & \textbf{Long-term Memory} \\
        \midrule
        GPT-4o (0806)  & 33.7 & 63.1 & 59.3 \\
        Gemini-Pro 1.5 & 31.7 & 55.4 & - \\
        Claude3-opus & - & 67.6 & -  \\
        Coze (GPT-4o) & 42.3 & 28.6 & 58.1 \\
        Kimi-Chat & 25.0 & 70.9 & - \\
        \midrule 
        \Name~(GPT-4o) & 39.4 & 37.2 & 59.0 \\
        \Name & 49.0 & 68.2 & 85.9 \\
    \bottomrule
\end{tabular}
\caption{The overall successful rates of different end-to-end systems.}
\label{table:overall_result}
    \vspace{1em}
\end{table*}

\subsection{Overall Results}
We present the overall results from our experiments in \autoref{table:overall_result}. We see that \Name~can achieve the best results on real-time information management and long-term memory and comparable performance with state-of-the-art systems in the management of private information.

For real-time information management evaluation, the primary limitation of the baseline systems is their inability to directly interact with target websites, preventing them from completing tasks such as \textit{adding items to cart} or \textit{rating a movie on IMDB} (except Coze, which possesses web-browsing plug-ins). Furthermore, our observations reveal a consistent pattern of inherent behavioral biases across the systems. For example, Gemini often sticks to Google-related products/websites while ignoring the instruction, e.g., searching for music playlists on YouTube despite the instruction asking for Soundcloud. Also, Kimi-Chat nearly answers all of the questions in Chinese and it can only access websites available in China, which leads to its low success rate. While an ideal system should not be impacted by such behavioral bias, we understand that the design decisions are also bounded by company and governmental policies.

In the private information management evaluation, all baseline systems can handle user-uploaded files, taking them as input to generate responses to user queries. Among the systems based on their own trained LLMs, such as GPT-4o, Gemini-Pro 1.5, and Claude3-opus, Coze stands out as a fully prompt-based system using GPT-4o as its backbone LLM. However, Coze and our Cognitive Kernel with GPT-4o performed poorly in handling user documents with a prompt-driven approach, as the system's instructions did not align well with the LLMs' training. Kimi-Chat outperformed all systems, particularly in managing long-context questions. GPT-4o ranks behind Kimi-Chat and Claude3-opus, often fabricating information not present in the document, especially when users ask unanswerable questions. Further analysis is provided in Section \ref{sec:private}.

For long-term memory management evaluation, we selected GPT and Coze as baseline methods, as they are the only two publicly available systems that support long-term memory.
As shown in \autoref{table:overall_result}, GPT-4o slightly outperforms Coze, indicating that its long-term memory module design is superior.
However, as discussed later in Section \ref{sec:memory_eval}, GPT-4o is susceptible to memory overwriting, which can result in catastrophic forgetting.
In contrast, \Name~achieves the best performance among all systems, with an overall accuracy of $85.9\%$.
Additionally, we evaluated the scenario where the base LLM in the Cognitive Kernel is replaced with GPT-4o.
The results showed a significant decrease in performance (from $85.9\%$ to $59.0\%$), as GPT-4o is not specifically aligned with our system design.

Next, we conduct an in-depth analysis of each core ability. 

\begin{figure}
    \centering
    \includegraphics[width=\linewidth]{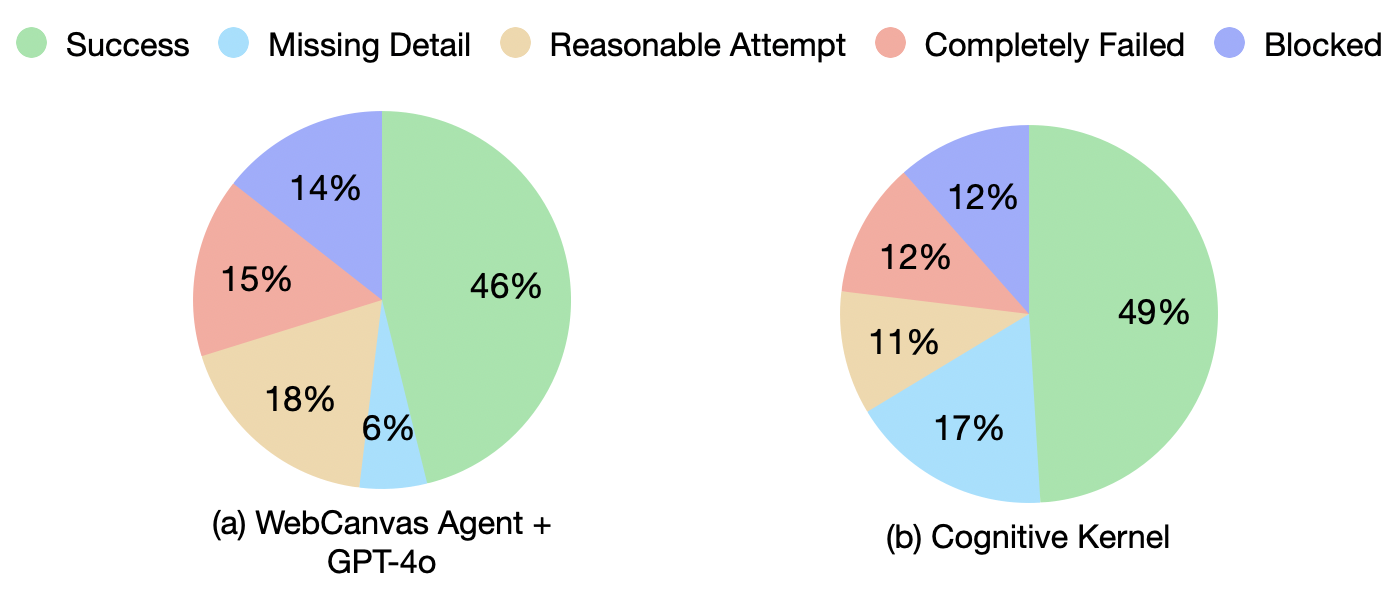}
    \caption{Overall task completion results on the WebCanvas test set.}
    \label{fig:web_res}
\end{figure}

\subsection{Real-time information management}
We further conduct experiments using a specialized web agent system specifically designed for web interaction. In particular, we rerun the WebCanvas agent~\citep{pan2024webcanvasbenchmarkingwebagents} using the GPT-4o API as the backbone and then manually evaluate its trajectories. We further categorized the failure cases for WebCanvas agent and \Name~to better understand the limitations of existing systems. 
Specifically, we define the following categories: \textbf{Missing Detail}: the system almost succeeded but missed a detail in the instruction. 
\textbf{Reasonable Attempt}: the system makes reasonable actions on the webpage (similar to a human when navigating an unseen website), but runs out of max steps before completing the task due to unfamiliarity. 
\textbf{Completely Failed}: the system's action trajectory does not make much sense. 
\textbf{Blocked}: the system's web browsing is blocked by bot detectors or Captcha. 

From \autoref{fig:web_res}, we see that the WebCanvas Agent system coupled with GPT-4o achieves a much higher success rate than baselines shown in \autoref{table:overall_result}, showing that the model capability is not the only factor that affects end-task performance. Also, in addition to the highest success rate, \Name~ also has a much larger portion of cases where the system almost completed the task, suggesting the overall superiority of the system. We showcase an example of \textbf{Missing Detail} in \autoref{fig:missing} and an example of \textbf{Reasonable Attempt} in \autoref{fig:reasonable} in the appendix.

\begin{figure*}[t]
\centering
{\includegraphics[width=0.47\textwidth]{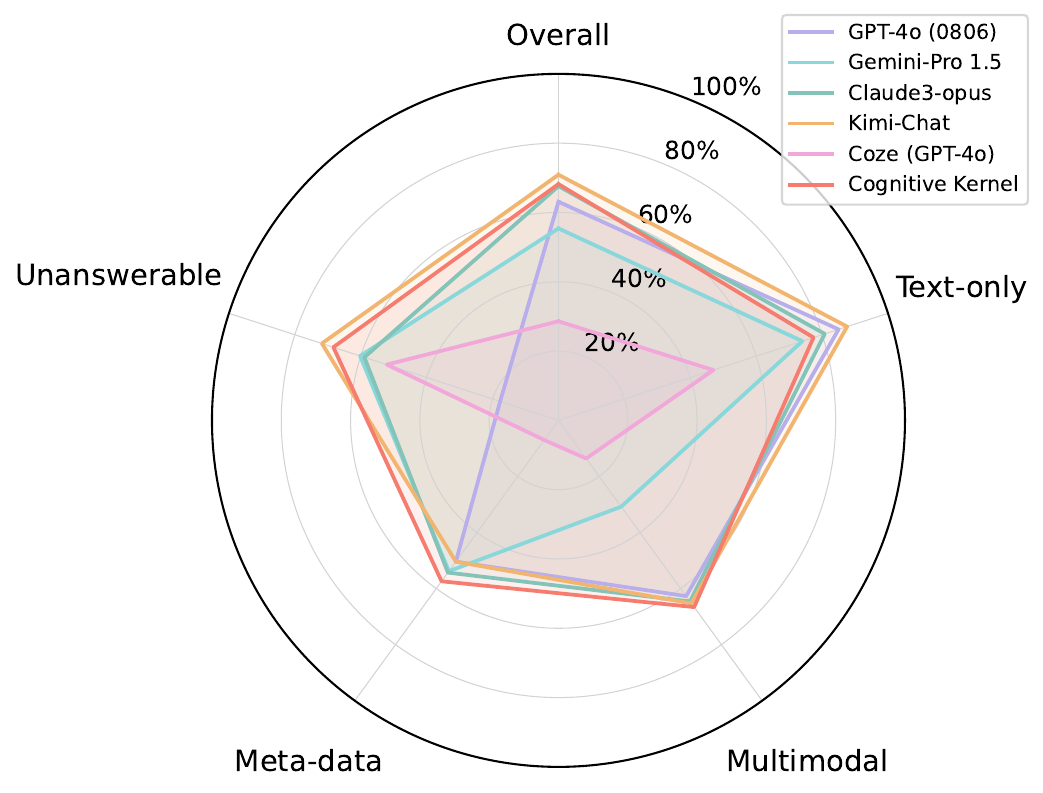}\label{fig:template1}}
{\includegraphics[width=0.49\textwidth]{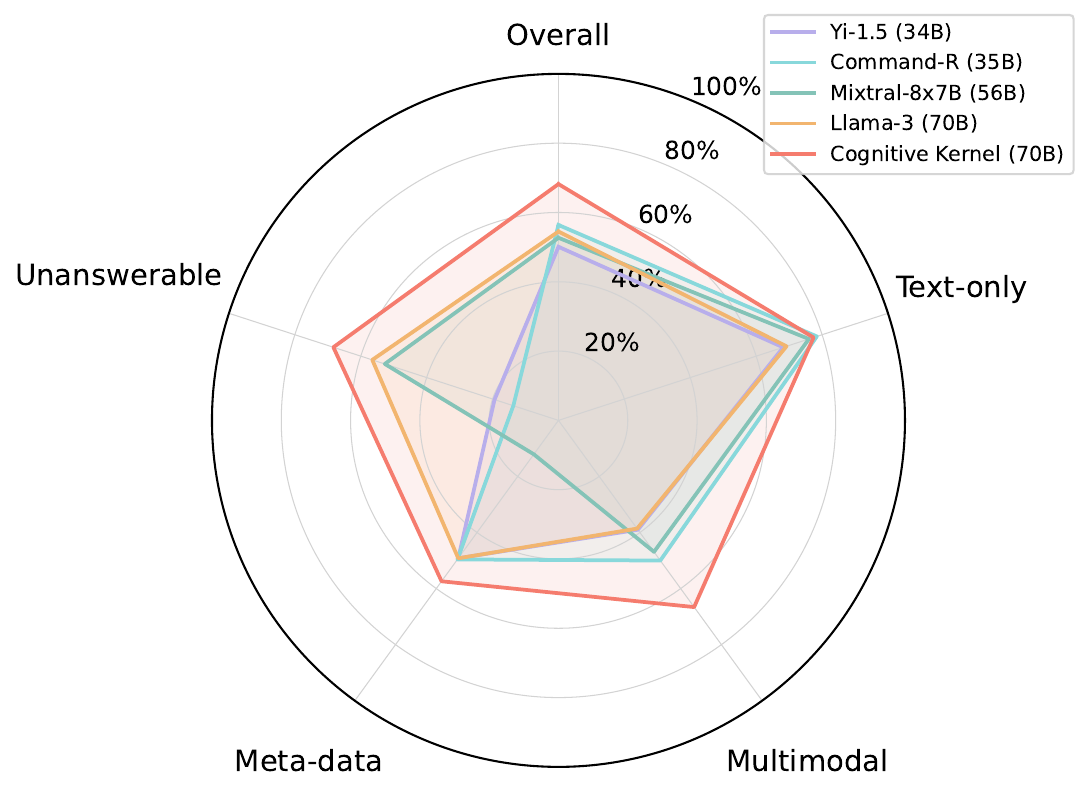}\label{fig:template1}}
\caption{Performance comparison of various end-to-end systems (left) and open-source LLMs across different question types in the DocBench.}
\label{fig:radar_doc}
\end{figure*}

\subsection{Private information management}
\label{sec:private}

\autoref{fig:radar_doc} presents a radar plot illustrating the accuracy of various end-to-end LLM-based systems and open-source LLMs using a \textit{parse-then-read pipelines} across different question types in the DocBench.
The questions in DocBench are categorized into four major types: Text-only, Multimodal (including tables and figures), Meta-data, and Unanswerable questions.

The left subfigure compares these systems with the end-to-end systems presented in Table \ref{table:overall_result}. Diving into this detailed comparison, we can observe the specific capabilities of these systems in handling different document-based questions. As shown, Kimi-Chat and Claude3-opus perform well across all these categories, demonstrating balanced performance on different types of questions. Notably, GPT-4 underperforms in the unanswerable category, suggesting potential overfitting in GPT-4's optimized file systems, likely due to training on datasets that only include answerable questions with provided golden answers. On the other hand, Gemini-Pro 1.5 struggles with figure-based questions in documents, with its performance in the multimodal category primarily driven by table-based questions. Coze (GPT-4o) performs poorly in handling user documents due to misalignment between system instructions and the LLM's instruction-following capabilities. Other systems show relatively balanced performance, with gaps mainly attributable to the backbone LLMs and their system designs.
As shown in the right subfigure, we also compare the results with recent state-of-the-art open-source LLMs using \textit{parse-then-read pipelines}. To enable LLMs to process documents as input, we use the \texttt{fitz} package to extract text, tables, and images from \texttt{PDF} files, then feed the questions, along with the extracted information, into the LLM to obtain the final answer. It is evident that simply using these LLMs does not lead to strong performance on document-based tasks. Compared to Llama-3 70B, \Name~demonstrates a 22\% relatively improved performance, highlighting the importance of system design in handling diverse types of user questions.





\subsection{Long-term memory management}
\label{sec:memory_eval}

    \begin{table*}[t]
	\centering
        \small
	\setlength{\tabcolsep}{4pt}
	\begin{tabular}{c|cccc|ccccc|c}
		\toprule
		\multirow{2}{*}{Systems} & \multicolumn{4}{c|}{Human-written test cases} & \multicolumn{5}{c|}{LLM-generated test cases} & \multirow{2}{*}{\textbf{Avg}} \\
            & 1-M & Mul-M & Update & All & 1-M & Mul-M & Update & Temp & All \\
            \midrule
            GPT-4o-mini & 70.0 & 70.0 & 80.0 & 73.3 & 100.0 & 71.4 & 68.0 & 66.7 & 75.9 & 74.6 \\
            GPT-4o & 60.0 & 50.0 & 70.0 & 60.0 & 76.0 & 40.5 & 84.0 & 45.8 & 58.6 & 59.3 \\
            Coze (GPT-3.5-turbo) & 60.0 & 50.0 & 60.0 & 56.7 & 48.0 & 9.5 & 36.0 & 8.3 & 23.3 & 40.0 \\
            Coze (GPT-4o) & 80.0 & 80.0 & 90.0 & 83.3 & 68.0 & 14.3 & 24.0 & 37.5 & 32.8 & 58.1 \\
            \midrule
            Cognitive Kernel (GPT-4o) & 70.0 & 50.0 & 50.0 & 56.7 & 68.0 & 73.8 & 40.0 & 54.2 & 61.2 & 59.0 \\
            Cognitive Kernel & 100.0 & 90.0 & 90.0 & 93.3 & 100.0 & 81.0 & 84.0 & 45.8 & 78.4 & 85.9 \\
		\bottomrule
	\end{tabular}
	\caption{The results of long-term memory management.}
	\label{table:memory_result}
        \vspace{1em}
    \end{table*}

    The detailed evaluation results of \Name, along with baseline systems on long-term memory, are presented in \autoref{table:memory_result}.
    While GPT-4o is generally more powerful than GPT-4o-mini, it performs significantly worse in long-term memory evaluation.
    Upon closely examining the intermediate content within the long-term memory module, we discovered that GPT-4o is more prone to modifying or overwriting existing memories when receiving new input from the user, even when the new content is only semantically similar to the old memory rather than actual update.
    As a result, the system may lose the ability to answer the final question accurately.
    Additionally, Coze does not perform as well as GPT in this setting.
    The choice of the underlying base model also has a significant impact on Coze's performance, with Coze + GPT-3.5-turbo performing substantially worse than Coze + GPT-4o. 
    In our system, we found that the Cognitive Kernel's performance is suboptimal when using GPT-4o as the base LLM.
    This is likely because GPT-4o is not fully aligned with our system prompt even though the prompt is natural and accurate for humans.
    After switching to the adapted LLM, the performance improved from $59.0\%$ to $85.9\%$.
    This observation shows that there is no perfect model and a continuously evolving AI system is crucial in real applications.

\section{Applications}\label{sec:applications}
\begin{figure}
    \centering
    \includegraphics[width=1.0\linewidth]{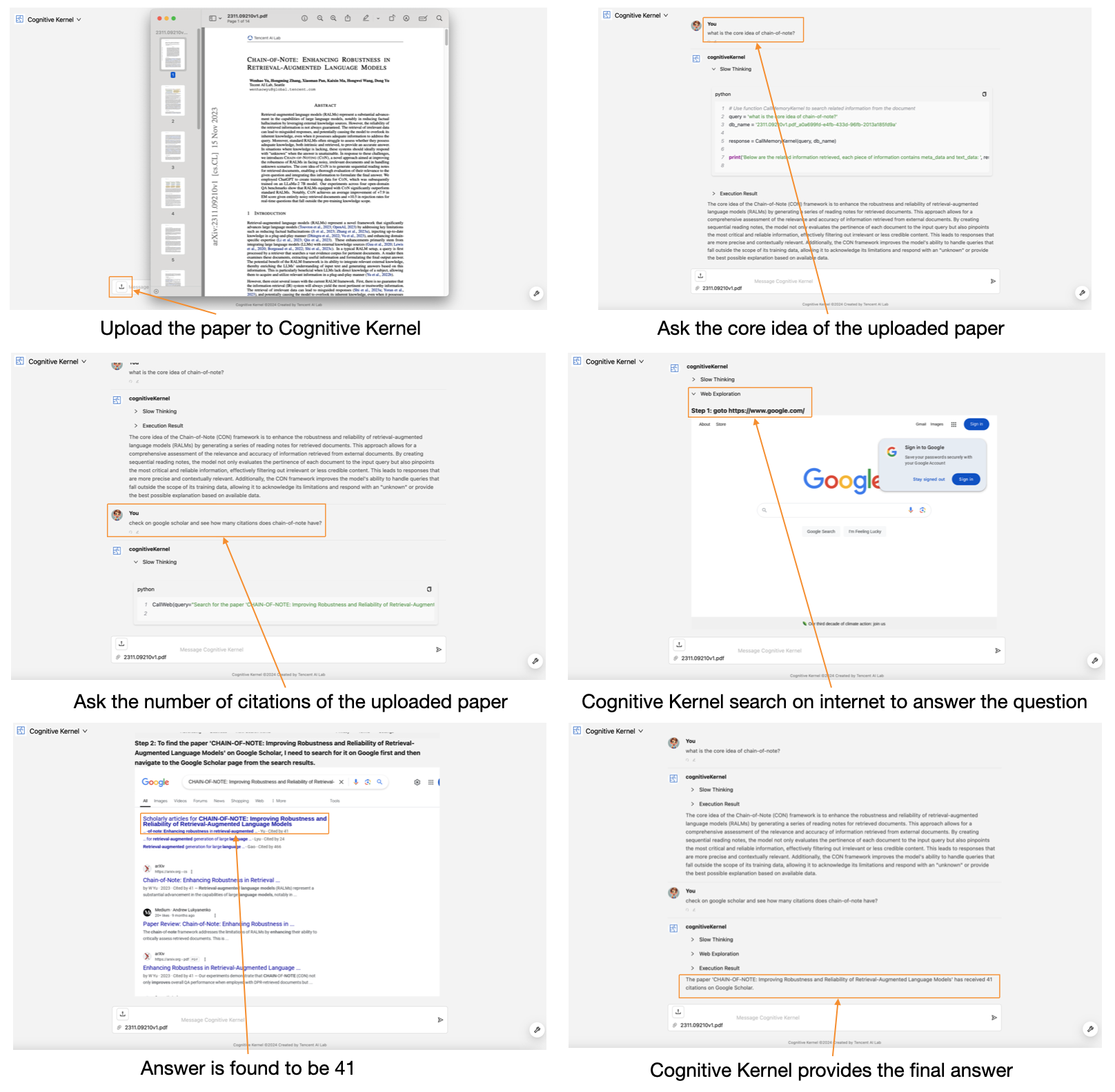}
    \caption{An example trajectory of \Name~ on completing user queries leveraging both private information management and real-time information management abilities.}
    \label{fig:case1}
\end{figure}

\begin{figure}
    \centering
    \includegraphics[width=1.0\linewidth]{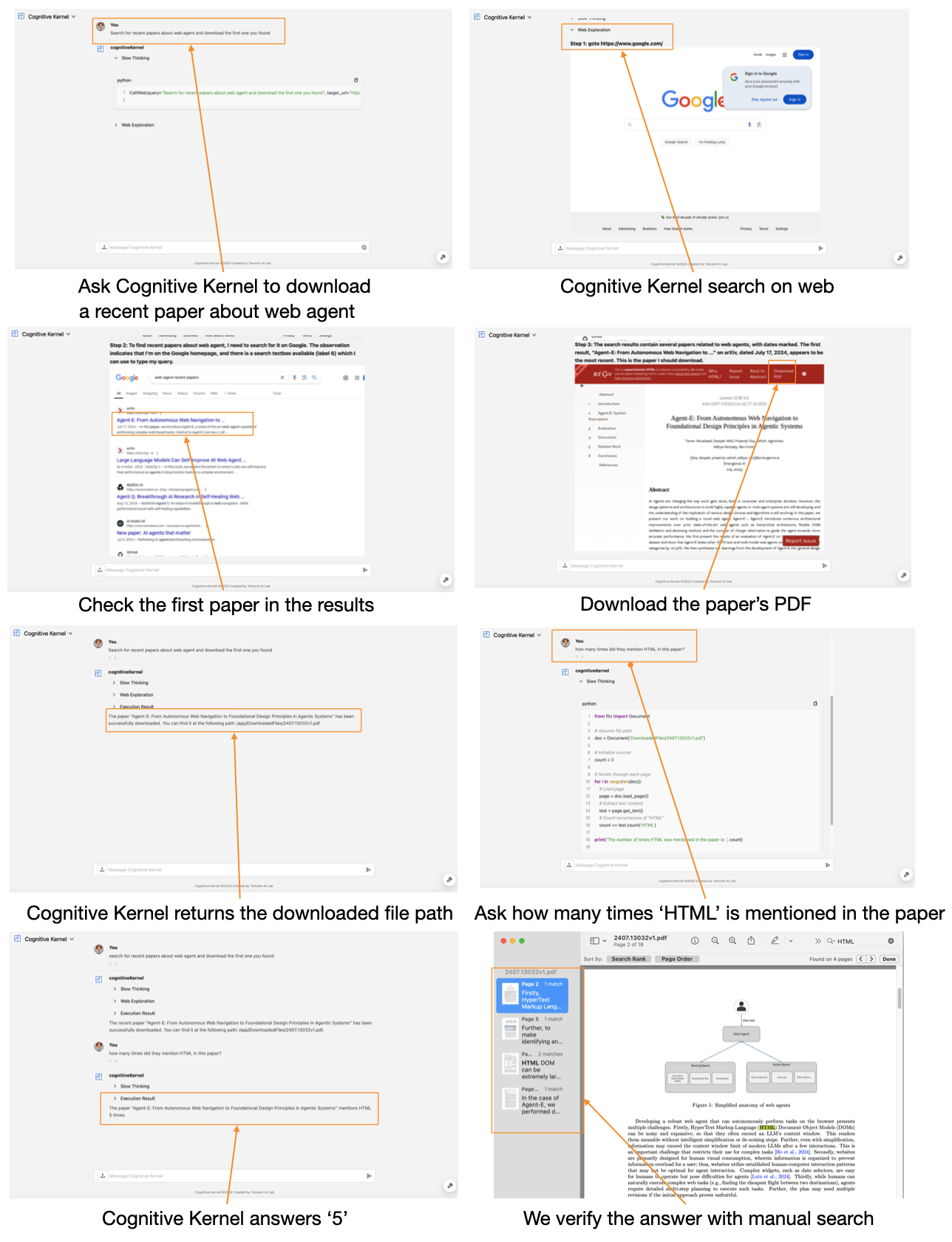}
    \caption{An example trajectory of \Name~ on completing user queries leveraging both real-time information management and private information management abilities.}
    \label{fig:case2}
\end{figure}

In this section, we showcase two application scenarios of \Name~to illustrate how users might interact with the system. The first example is shown in \autoref{fig:case1}. In this case, the user first uploaded a paper called ``Chain-of-note''\citep{yu2023chainofnote} to the system and asked about the paper's core idea. \Name~first processed and indexed the uploaded document internally, and then read the document to answer the question. In the next turn, the user asked about the current number of citations of Chain-of-note, and the system recognized this question to be a real-time information-seeking one and instantiated a web browser to look for the evidence. Finally, based on the search results, \Name~provided the correct answer.  In the second example from \autoref{fig:case2}, we see that the user first asked the system to search for recent papers about web agents and download the first one it found. \Name~again opened a web browser and searched web agent papers. Then it opened the first paper in the results and clicked the download button on the paper's arXiv page. After the paper was downloaded, the system returned the downloaded file's path. In the next turn, the user then asked how many times the keyword ``HTML'' is mentioned in the paper. Then \Name~leveraged its private information management ability to open the paper and count the occurrences of ``HTML.'' Then it returned the answer \textit{5}, which we verified to be correct by opening the paper and manually searching the keyword.

\section{Discussions and Limitations}

Despite that \Name~achieves promising performance on several realist tasks, there is still a huge gap between \Name~and a generalist ``autopilot'' system.
In this section, we discuss the limitations of our current system, which is also our future working directions.

\subsection{Multi-modal Perception Ability}
For a generalist ``Autopilot'' system, it's critical to have multi-modal perception ability because the real world is multi-modal and the rich information that helps decision-making is often embedded in other modalities beyond text. 
However, current \Name~employs an LLM as the central policy model, thus it cannot handle multi-modal inputs such as images or audio. 
Intuitively, all three use cases we experimented with could greatly benefit from other modalities and provide a better user experience. 
For example, for private information management, the system can better understand local files by reading both the text and images embedded in the files. For real-time information management, the system can better understand the websites by observing the visual layout and visual elements not captured in the accessibility tree, therefore navigating the web more effectively. 
For long-term memory management, the user can also send images to the system or speak to it directly without typing any text, and the system can read and write memories of different modalities to better serve different scenarios. 
Therefore it is a promising direction to equip \Name~with multi-modal perception ability and we leave this exploration for future work. 

\subsection{Self-improvement Through Search and Feedback}
Even though we give \Name~freedom to be a generalist autopilot system, it tends to mimic the training trajectories.
Thus, an important research question is how to make it capable of continuously evolving to overcome the limitation of the limited training trajectories such that it could generalize to unseen tasks.
Similar to previous efforts on reinforcement learning, a promising way towards this goal is by letting the system search for different task-solving strategies in unseen situations, automatically collect feedback signals for different trajectories, and orchestrate the learning and system updates autonomously. 
The most critical component in this pipeline is the feedback signal.
Currently, \Name~still relies on users or external models to provide the feedback, which limits its scalability. 
Ideally, the system should also acquire the ability to be a critic, such that it can serve as a value function to estimate the quality of its own task-solving trajectories. 
With such ability, the system could adopt algorithms such as Monte-Carlo Tree Search (MCTS) \citep{tian2024selfimprovement} to explore the real-world environments, collect the self-feedback, and continue improving itself with this signal. 
We believe this direction is a necessary path toward a true ``Autopilot'' system and warrant further investigation. 

\subsection{Robust System-level Support}
Since the purpose of an ``Autopilot'' system is to complete various user tasks in the real world, it requires much more than a powerful policy model. 
In other words, the system needs a robust and scalable infrastructure to support the central policy model to function as the ``brain'' an intelligent agent, much like the relationship between the human body and the brain. 
With this in mind, we adopted a dockerized design and equipped \Name~with various functionalities such as file processing and web interaction, as described in Section \ref{sec:system_design}. 
However, we recognize that what we currently have is still quite limited and there is ample room for improvement in terms of both scope and robustness. 
For example, our system only supports user-uploaded file processing for private information management. 
To further expand the scope of applications, we could allow the system to directly access the local file systems and control other software on the operating system level \citep{xie2024osworldbenchmarkingmultimodalagents,wu2024oscopilot}. 
In this way, we can enable the system to manage different kinds of private information for the user. 

From the perspective of robustness, we also noticed that our system could not cover all edge cases in the environment. For example, when \Name~leverages the browser to complete web-based tasks. In certain cases, despite the policy model predicting the correct action, the action could not be executed in the web server due to certain technical issues (e.g., elements in iFrames could not be directly clicked as other buttons). 
Concurrent works on web agents have also observed that such technical issues contribute to a large portion of failure cases \citep{yoran2024assistantbench}. 
We believe that improving the system's robustness is equally important as improving the capabilities of the policy model. Only the organic combination of these two can truly unleash the power of large language models and build ``Autopilot'' systems. 
\section{Conclusion}
\label{sec:conlusion}

In this technical report, we presented \Name, a dockerized agent system towards the goal of general-purpose ``Autopilots.'' 
Inspired by the idea of the Turing machine, we formulate two fundamental duties of an ``Autopilot'' system: state management and decision-making. Based on this formulation, we proposed a system architecture with three main components: reasoning kernel for decision-making, perception kernel for state perceiving, and memory kernel for state management. For implementation, we further reorganized these components into several dockers to pursue easy and safe deployment. We conduct experiments to evaluate \Name's ability to handle user requests in three scenarios: real-time information management, private information management, and long-term memory management. The evaluation results show that \Name~achieves stronger or similar performance compared to other SOTA closed-source systems. We release our system framework, model weights, and findings to the community, hoping to inspire future research on the direction of generalist ``Autopilots'' systems. 
\section*{Disclaimer}

Since \Name~directly interacts with the real world (through the web browser and host file system\footnote{By default, \Name's dockers are isolated such that they could not impact the host system. However, one could bind the host system's file path with docker's file path, e.g. for debugging purposes, which would make it possible for \Name~to directly access the host file system.}), it might introduce additional risks for the user. For example, accessing confidential information on the host system or downloading malicious content from the internet. Thus it requires extra caution and a substantial amount of safe checks when deploying and running the \Name. In our experiments that involve the open web, we also closely monitor the \Name~'s progress to make sure that the system is not causing any harm to the website hosts. 

Also, we would like to note that \Name~is a research project and not an official product from Tencent. Our intention is to explore the possibility of building ``Autopilot'' systems and we hope our work could inspire others and bring more positive social impact. Any misuse of the system that could be potentially harmful to others is strictly forbidden. 

\section*{Acknowledgement}
We would like to thank Sihao Chen, Tong Chen, Tianqing Fang, Hongliang He, Ruixing Hong, Zhehui Huang, Mengzhao Jia, Eric Lan, Siru Ouyang, Dan Roth, Zhaowei Wang, Di Wu, Sherry Wu, Zilin Xiao, Yuwei Zhang, Zhisong Zhang, Xinran Zhao, Ben Zhou, and Anni Zou for their valuable help and suggestions for building \Name. We also want to thank all members of the Tencent AI Lab for their insightful comments and feedback on our work. 

\newpage
\bibliographystyle{tencent_ailab_tech_report}
\bibliography{tencent_ailab_tech_report}

\newpage
\section*{Appendix}
\subsection{Trainng Details}
\label{app:training}
\paragraph{Pre-trained Model} We train our own policy models based on Llama3 series~\citep{dubey2024llama}
\paragraph{Hyper-parameters} The hyper-parameters for learning the 70B and 8B policy models are listed in \autoref{tab:heyper-param}.
\begin{table}[!htb]
\small
    \centering
    \begin{tabular}{l|ccccc}
    \toprule
    \multicolumn{1}{c}{}     &  Learning Rate & Sequence Length & Warmup Step & Global Batch Size & Epoch \\
    \midrule
     70B     & 1e$-5$  &  8192 & 0 & 64 & 3  \\
     8B      & 5e$-5$  & 8192  & 50  & 64  & 3 \\ 
    \bottomrule
    \end{tabular}
    \caption{Hyper-parameters for the 70B and 8B policy models.} 
    \label{tab:heyper-param}
\end{table}
\paragraph{Training Data}
\autoref{tab:data} displays the training data statistics for the proxy model, primarily divided into conventional instruction-following data and agent data.
\begin{table}[!htb]
\small
    \centering
    \begin{tabular}{c|ccccc}
    \toprule
    \multicolumn{1}{c}{}     &  \# Samples  \\
    \midrule
     Instruction-following     & 33,882 \\
     Agent                     & 49,996 \\
    \midrule
     Total                     & 83,878 \\
    \bottomrule
    \end{tabular}
    \caption{Training data statistics for the policy model.} 
    \label{tab:data}
\end{table}

\subsection{Frontend UI}
\label{app:frontend}
\begin{figure}
    \centering
    \includegraphics[width=1.0\linewidth]{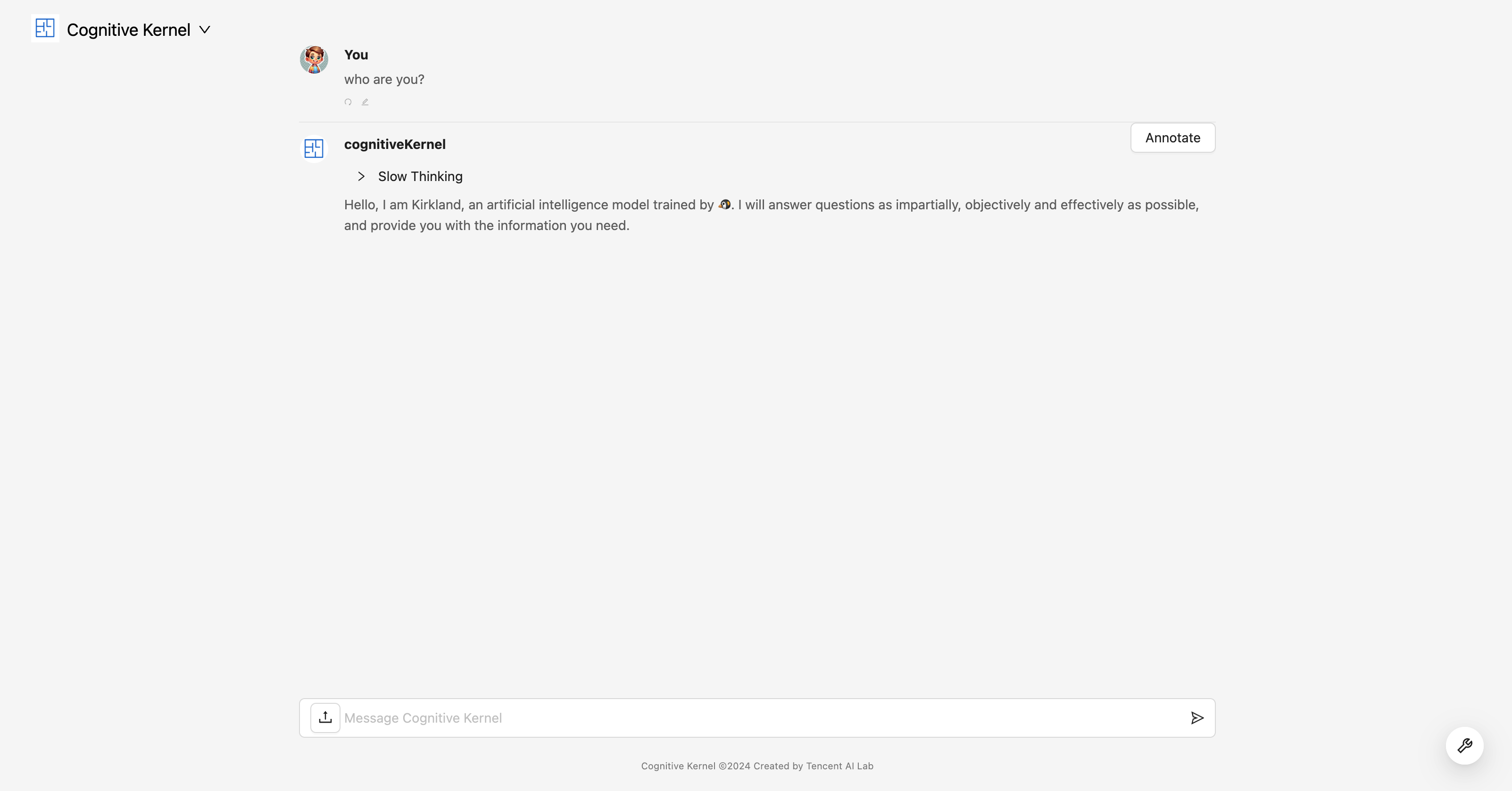}
    \caption{\Name~'s frontend user interface}
    \label{fig:UI}
\end{figure}

\begin{figure}
    \centering
    \includegraphics[width=1.0\linewidth]{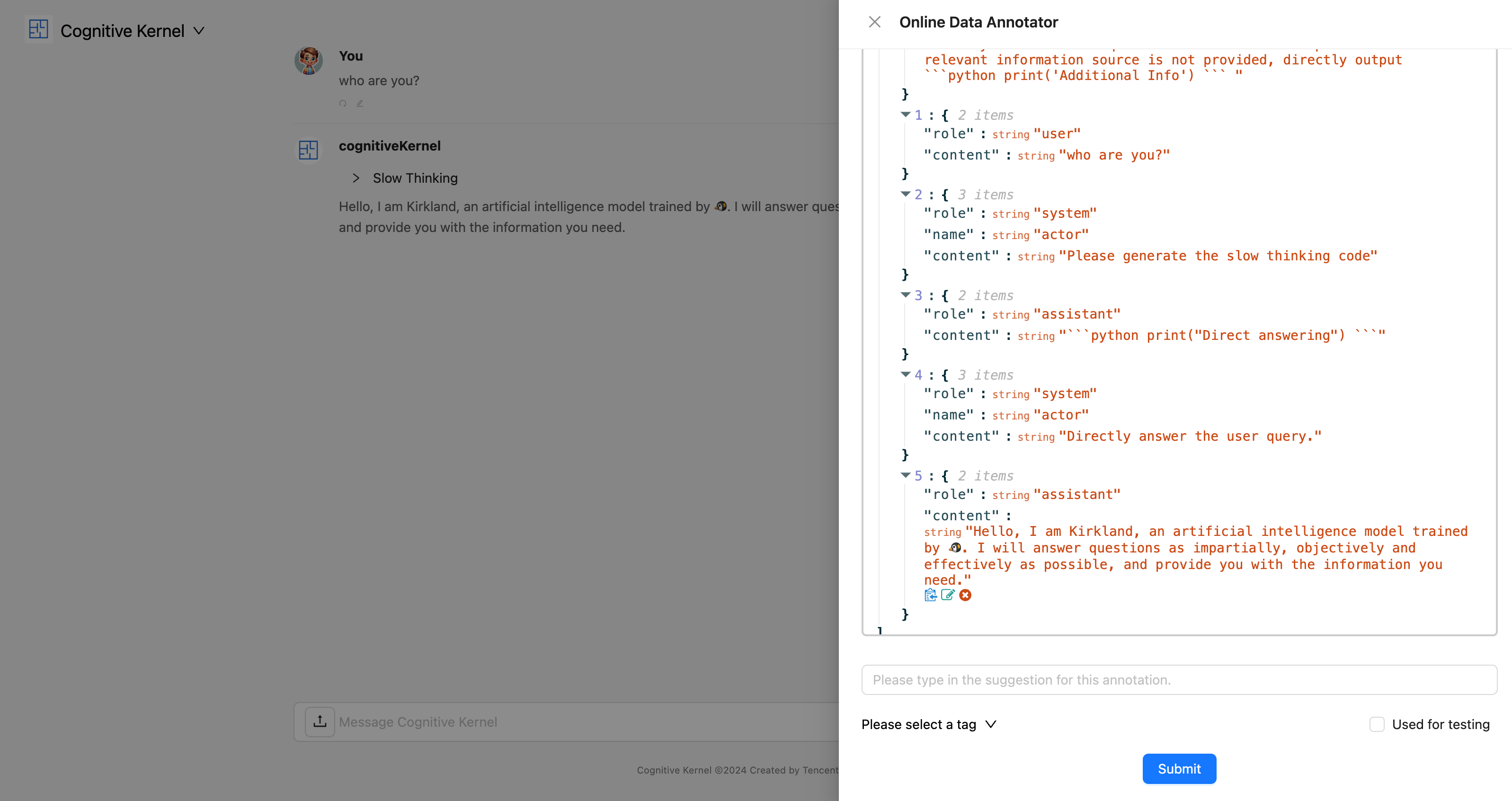}
    \caption{\Name~'s frontend user interface when activating the feedback mode}
    \label{fig:annotation}
\end{figure}

In \autoref{fig:UI}, we show an illustration of the frontend UI of \Name. The user can optionally activate the annotation mode of the system, after which an \textit{Annotate} button will appear right next to each turn of the dialog. Upon clicking the button, the annotation interface will show up, as illustrated in \autoref{fig:annotation}. This interface shows the full prompt of the current dialog session in the message format, including the system prompt turns. The user has the freedom to directly edit the assistant response and submit the modified message to a persistent database. Besides editing, the user also has the option to provide suggestions for edits, and all of the data and changes are saved in the database. This interface can be handy for collecting user feedback data, which is critical for continue improving the system. 

\subsection{Error Cases}
\begin{figure}
    \centering
    \includegraphics[width=1.0\linewidth]{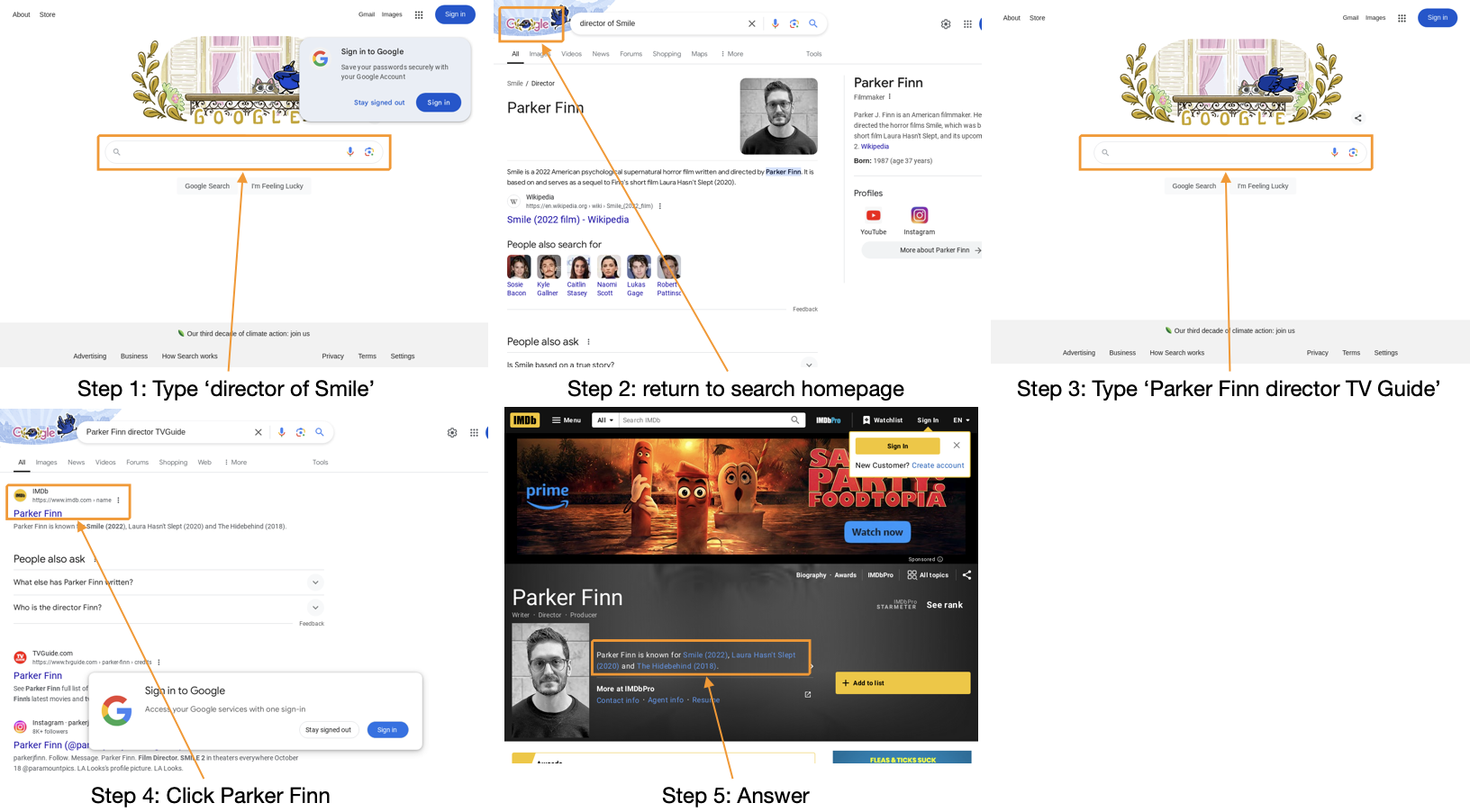}
    \caption{An example trajectory of \textbf{Missing Details} from \Name. The task instruction is "\textit{Find more films from the director of Smile on tvguide}"}
    \label{fig:missing}
\end{figure}

In \autoref{fig:missing}, we illustrate an example of \textbf{Missing Details} from the WebCanvas test set. Given the query, the system first searched for the director of Smile and found it to be Parker Finn. Then it tried to search for other movies directed by Parker Finn, and it found the answer from the IMDB website. However, the initial query requires the answer from the TVGuide website, thus the task is not considered as successful. 

\begin{figure}
    \centering
    \includegraphics[width=1.0\linewidth]{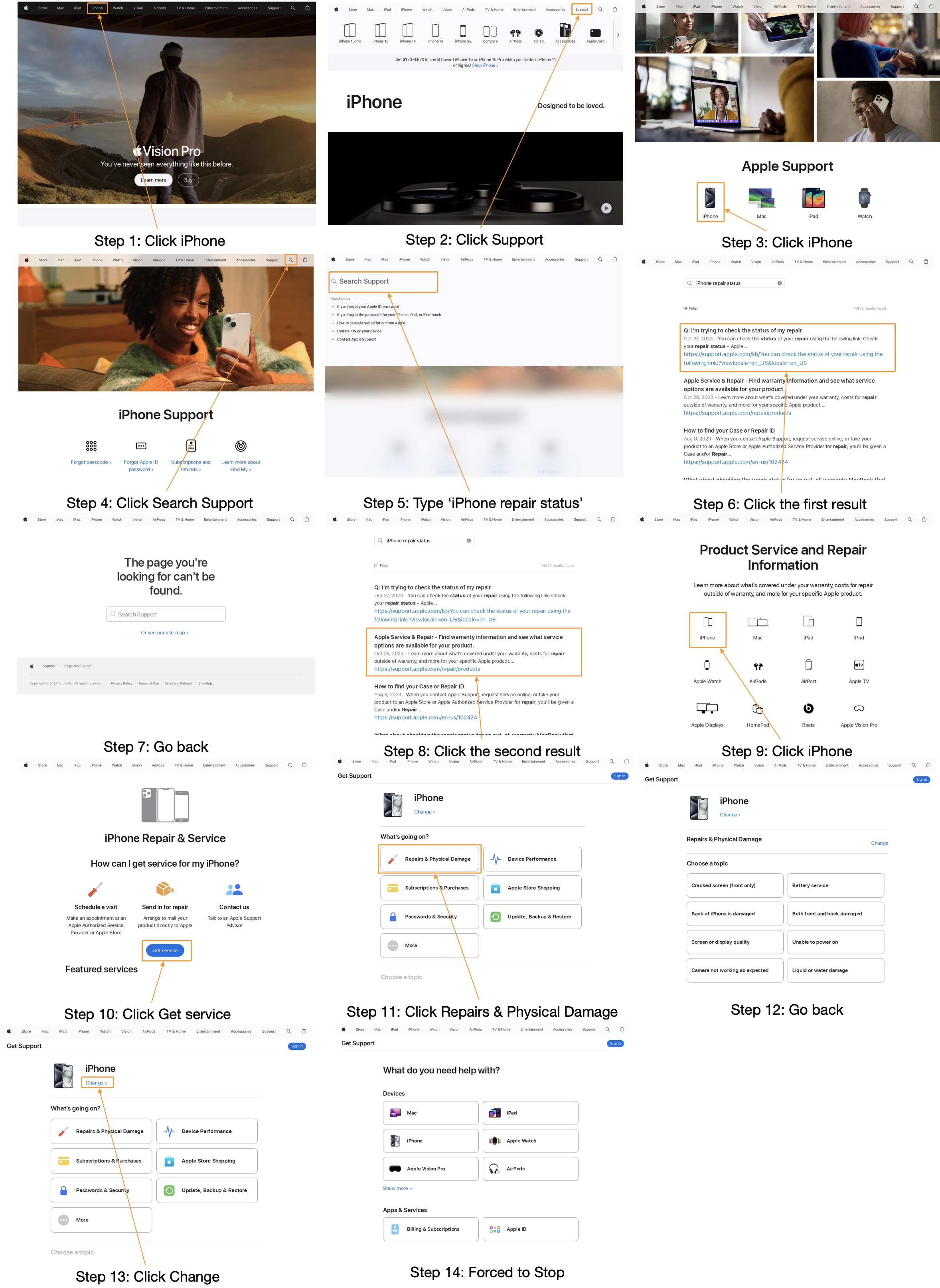}
    \caption{An example trajectory of \textbf{Reasonable Attempt} from \Name. The task instruction is "\textit{Check the status of your iPhone repair on apple.}"}
    \label{fig:reasonable}
\end{figure}

In \autoref{fig:reasonable}, we show an example of \textbf{Reasonable Attempt}. The task requires checking the iPhone repair status, and it would be considered successful if it can reach the Apple My Support page \footnote{\url{https://support.apple.com/my-support}} (upon signing in one can check the status there). In this case, \Name~ examined different pages on the Apple website that are relevant to iPhone repair but could not find the page with explicit information for the status check. After reaching the maximum number of steps it is forced to stop. We believe such errors are reasonable and could be potentially solved by allowing the system to explore the website before executing any tasks and update its memory accordingly. 

\end{document}